\documentclass[conference]{IEEEtran}
\IEEEoverridecommandlockouts
\usepackage[colorlinks,urlcolor=blue,linkcolor=blue,citecolor=blue]{hyperref}
\usepackage{times}
\usepackage{epsfig}
\usepackage{graphicx}
\graphicspath{ {./Fig/} }
\usepackage{amsmath}
\usepackage{amssymb}
\usepackage{pifont}
\newcommand{\cmark}{\ding{51}}%
\newcommand{\xmark}{\ding{55}}%
\usepackage{commath}
\usepackage{rotating}
\usepackage{amsthm}
\usepackage{algorithmic}
\usepackage[super]{nth}
\usepackage{booktabs}
\usepackage[norelsize,longend,ruled,lined,boxed,commentsnumbered]{algorithm2e}
\usepackage{breqn}
\usepackage{multirow}
\usepackage{caption}
\usepackage{subcaption}

\begin{document}

\title{3D-Convolution Guided Spectral-Spatial Transformer for Hyperspectral Image Classification

\thanks{\noindent$^{\ast}$Worked as Interns.\\
\\
This paper is accepted in IEEE Conference on Artificial Intelligence, 2024. IEEE copyright notice \copyright 2024 IEEE. Personal use of this material is permitted. Permission from IEEE must be obtained for all other uses, in any current or future media, including reprinting/republishing this material for advertising or promotional purposes, creating new collective works, for resale or redistribution to servers or lists, or reuse of any copyrighted component of this work in other works.}

\author{Shyam Varahagiri$^{\ast}$, Aryaman Sinha$^{\ast}$, Shiv Ram Dubey, Satish Kumar Singh\\
Indian Institute of Information Technology, Allahabad, India\\
{\tt\small \{varahagiri.shyam, aryaman12483sinha\}@gmail.com, \{srdubey, sk.singh\}@iiita.ac.in}
}
}

\maketitle

\begin{abstract}
In recent years, Vision Transformers (ViTs) have shown promising classification performance over Convolutional Neural Networks (CNNs) due to their self-attention mechanism. Many researchers have incorporated ViTs for Hyperspectral Image (HSI) classification. HSIs are characterised by narrow contiguous spectral bands, providing rich spectral data. Although ViTs excel with sequential data, they cannot extract spectral-spatial information like CNNs. Furthermore, to have high classification performance, there should be a strong interaction between the HSI token and the class (CLS) token. To solve these issues, we propose a 3D-Convolution guided Spectral-Spatial Transformer (3D-ConvSST) for HSI classification that utilizes a 3D-Convolution Guided Residual Module (CGRM) in-between encoders to ``fuse" the local spatial and spectral information and to enhance the feature propagation. Furthermore, we forego the class token and instead apply Global Average Pooling, which effectively encodes more discriminative and pertinent high-level features for classification. Extensive experiments have been conducted on three public HSI datasets to show the superiority of the proposed model over state-of-the-art traditional, convolutional, and Transformer models. 
The code is available at \href{https://github.com/ShyamVarahagiri/3D-ConvSST}{https://github.com/ShyamVarahagiri/3D-ConvSST}.
\end{abstract}


\section{Introduction}
\IEEEPARstart{H}{yperspectral} images (HSIs) contain several spectral bands at each pixel, which enables the recognition of materials. Particularly those that have minute spectral discrepancies \cite{nonconvexmodel} can be easily identified. In addition, HSIs also contain 2D spatial information, which is essential for improving the representation of hyperspectral data 
 \cite{frontiersspatialspectral}. Due to the rich spectral and spatial information possessed, HSIs have been employed for earth observation and remote sensing tasks, like urban planning \cite{hsiurban}. crop management \cite{hsiagri}, and environmental monitoring \cite{hsienvironment}.

Classical HSI classification techniques involved pixel-wise classification of the spectral signatures (e.g., support vector machines (SVMs) \cite{hsisvm} and random forest (RF) \cite{randomforesthsi}). 
Since spatial contexts are not in consideration, classification maps produced by these pixel-wise classifiers are frequently inadequate \cite{hsioverview}. Due to poor representation capabilities and limited data fitting, traditional methods encounter performance bottlenecks as the training data becomes more complex.

For HSI classification, deep learning (DL) methods are prominent \cite{overviewDL}. 
HSI classification has been dominated by convolutional neural network (CNN) based architectures \cite{cnn}. CNNs apply multiple linear transformations combined with non-linear activation functions to extract spectral-spatial information. 
A 1D-CNN is used to extract the spatial features of HSIs in \cite{cnn1d}. A 2D-CNN is leveraged to encode the spatial characteristics of HSIs in \cite{cnn2d}. Dual branches are employed in \cite{branch} and \cite{branch1} where the spectral domain is captured using a 1D-CNN and the spatial domain is captured using a 2D-CNN. The two features are then fused using multiple fully connected (FC) layer and passed to classifier. A 3D-CNN is employed to capture the joint spatial-spectral features from HSIs in \cite{cnn3d}. HSI-CNN \cite{luo2018hsi} also utilizes a 3D-CNN to extract spectral-spatial features and then uses a standard 2D-CNN to extract the HSI features. In \cite{roy2019hybridsn}, a hybrid spectral-spatial CNN (HybridSN) uses a 3D-CNN to learn the joint spatial-spectral feature representation followed by the encoding of spatial features using a 2D-CNN. 
To extract the discriminative characteristics using CNNs, different methods are proposed such as integration of additional channel and spatial attention layers \cite{zhong2017spectral} and utilization of pyramidal ResNets \cite{paoletti2018deep}.

\begin{figure*}[!t]
\label{fig:architecture}
    \centering
    \includegraphics[width=0.79\textwidth]{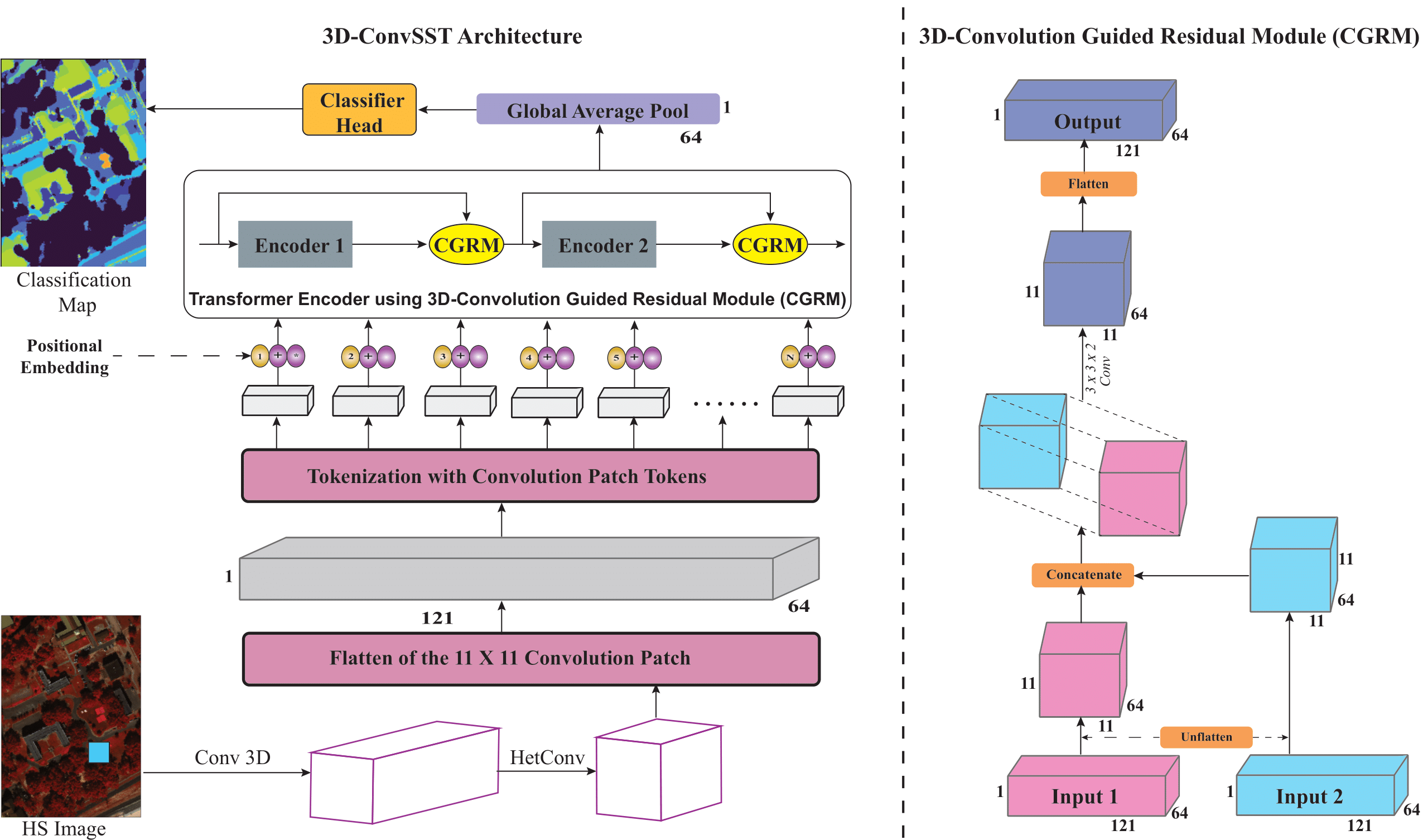}
    \caption{Overall framework of the 3D-ConvSST Architecture (left) and the 3D-Convolution Guided Residual Module (right)}
    \label{fig:SST_Architecture}
\end{figure*}

In recent years, Transformers are cutting-edge backbone networks that excel at processing sequential data by using a self-attention mechanism. Transformer networks have shown very promising performance for HSI classification. 
In \cite{he2021spatial}, a spatial-spectral Transformer (SST) network is exploited to construct connections between adjacent spectra using a modified transformer with a dense connection. 
SATNet uses the self-attention and spectral attention mechanisms to extract features from HSIs \cite{he2021spatial}.
The Vision Transformer (ViT) \cite{dosovitskiy2020image} has proven highly effective at processing visual data. It can effectively capture the relationship between distant HSI spectral patterns. CNN and ViT are combined in the recent HSI classification techniques. The Convolution Transformer Mixer (CTMixer) \cite{zhang2022convolution} combines the ability of CNNs to extract local features, and the ability of ViTs to extract global features by using a group parallel residual block and a dual-branch structure. In \cite{sun2022spectral}, a gaussian-weighted feature tokenizer Transformer is used to obtain deep semantic features. A group-aware hierarchical Transformer overcomes over-dispersion of the multiple HSI bands by using a grouped pixel embedding module \cite{mei2022hyperspectral}.
SpectralFormer \cite{hong2021spectralformer} uses group-wise spectral embeddings to learn spectral information. 
Best architecture configurations of spectral-spatial Transformer are searched for HSI classification in \cite{zhong2021spectral} using a factorized architecture search framework. 
Spectral-spatial transformer not only show better results, but is also less noise sensitive \cite{morphformer}.

The Transformer models mentioned above exploit the Transformer's ability to process spectral data. Although Transformers excel at capturing sequential spectral data, CNNs are effective at capturing high-level local features. In this paper, we propose a 3D-Convolution guided Spatial-Spectral Transformer (3D-ConvSST). This model utilizes the Transformer's ability to capture global information and spectral features and also utilizes a 3D-Convolution layer to extract local spatial features. In the proposed model, a 3D-Convolution layer followed by a HetConv \cite{hetconv} layer is used to introduce some inductive bias inherent to CNNs to the Transformer. The CNNs are also used to control the depth of the spectral features and to extract robust and discriminative features from the HSI. Inside the Transformer, a 3D-Convolution is used to fuse information from the preceding encoder blocks, before being fed to the next encoder block. This helps in the extraction of valuable spatial features. Finally, our model foregoes the usage of the class (CLS) token. Rather, we use global average pooling of the final features corresponding to each patch. Global average pooling takes advantage of the spatial data extracted by the 3D-Convolution. At the end, a linear transformation is utilized followed by softmax to produce the class-wise probabilities.
The main contributions are summarized as:
\begin{enumerate}
  \item A novel 3D-Convolution Guided Residual Module (CGRM) is proposed to fuse spectral information and extract spatial information from the preceding Transformer encoder blocks. It utilizes a 3D-Convolution layer between each encoder block.
  \item A global average pooling layer is used rather than a class token for final representation. This takes advantage of the spatial information extracted from the CGRM module to classify HSI images.
  \item A 3D-ConvSST Transformer model is introduced by using the CGRM module with global average pooling. The evaluations are performed on three public and benchmark HSI datasets to validate the efficacy.
\end{enumerate}

Rest of this paper contains the architecture of the 3D-ConvSST in Section \hyperref[sec:proposed]{II}, experimental settings in Section \hyperref[sec:experimental_settings]{III}, results in Section \hyperref[sec:experimental]{IV}, and conclusion in Section \hyperref[sec:conclusion]{V}.

\section{Proposed 3D-ConvSST Architecture} 
\label{sec:proposed}
Fig. \ref{fig:SST_Architecture} illustrates the proposed 3D-ConvSST architecture for HSI classification. It consists of the convolution-based feature tokenization, transformer encoder, 3D-convolution guided residual module, global average pooling and classifier.

\subsection{Convolutional Networks for Feature Extraction}
The proposed model incorporates CNN-specific inductive biases into the Transformer and extracts high-level abstract features using a convolutional neural network (CNN). This also allows us to control the spectral dimensions of the HSI. Robust and discriminative characteristics are extracted from the HSI by the suggested model using a \texttt{3D-Conv} layer followed by a \texttt{HetConv} \cite{hetconv} layer. The HSIs are cubic with shape $(H \times W \times B)$. It is reshaped into $(1 \times H \times W \times B)$ and fed into a \texttt{3D-Conv} layer with a kernel having $(3 \times 3 \times 3)$ dimension. This fully utilises the spectral and spatial information of the HSI. Padding of $(1 \times 1 \times 0)$ is used to keep the spatial dimensions consistent. The data is then reshaped and fed into the \texttt{HetConv} layer, which consists of two 2D-Convolutions working in parallel, one groupwise and one pointwise. \texttt{HetConv} utilizes two kernels to extract multiscale information. The results of the two operations are added elementwise ($\oplus$). Batch Normalization \cite{batchnormalization} and the ReLU activation function is applied after each convolution. 
These operations are represented as,
\begin{equation}
\begin{aligned}
         X &= Reshape(X_{HSI}) \\
         X_{i} &= Reshape(Conv3D(X, k1, p1, g1)) \\
         X_{o} &= HetConv(X_{i}) \\ 
         &= Conv2D(X_{i}, k2, p2, g2) \oplus Conv2D(X_{i}, k3, p3, g3), \\
\end{aligned}
\end{equation}
where $k1 = 3$, $p1 = (1, 1, 0)$, $g1 = 1$, $k2 = 3$, $p2 = 1$, $g2 = 8$, $k3 = 1$, $p3 = 0$, and $g3 = 1$. K is the kernel size of the convolution, p is the padding and g is the number of groups. The output shape of the \texttt{3D-Conv} is $(8 \times H \times W \times (B - 8))$ and that of the \texttt{HetConv} is $(H \times W \times 64)$.

\subsection{Tokenization and Positional Embedding}
The HSI cube must be embedded into patches before being fed into the Transformer. Multiple patch tokens having shape $(1 \times 64)$ can be obtained by flattening HSI subcubes of shape $((H \times W) \times 64)$ as $X_{flat} = T(Flatten(X_{o}))$, where T($\cdot$) is a transpose function and $X_{flat}$ $\in$ $\mathbb{R}^{n \times 64}$, and $n$ is the number of patches.
The sequential information of the patch tokens are retained using trainable positional embeddings (\textit{PE}) which is added to the HSI tokens, followed by a dropout layer as $X = DP(X_{flat} \oplus PE)$, where \textit{DP} represents a dropout layer with value $0.1$.

\subsection{Vision Transformer Encoder Module}
Vision Transformer (ViT) \cite{visionTransformer}  consists of a transformer encoder module that receives the features of different patches as input tokens and transforms them into output tokens. 
A multilayer perceptron (MLP) block and a multi-head self-attention (MSA) block comprise the Transformer encoder. A Layer Norm (LN) is used prior to the MSA and MLP blocks, with residual connections following each block. The Gaussian Error Linear Unit (GELU) is the activation function utilised for the MLP. The encoder block is represented as: 
\begin{equation}
    \begin{aligned}
        z'_{l} &= MSA(LN(z_{l-1})) + z_{l-1} \\
        z_{l} &= MLP(LN(z'_{l})) + z'_{l},
    \end{aligned}
\end{equation}
where $z$ is the feature tokens and $l = 1, 2, ..., L$. The performance of the Transformer can be attributed to its self-attention mechanism in the MSA block which captures the correlation between sequential tokens. This is achieved by linearly mapping the tokens to a Query \textbf{Q}, Key \textbf{K}, and Value \textbf{V} vectors. The attention score measures the strength of the relationship of a token with other tokens in the sequence. The self-attention is computed as
\begin{equation}
    Attention(\textbf{Q}, \textbf{K}, \textbf{V}) = softmax\left(\frac{\textbf{Q}\textbf{K}^T}{\sqrt{d_{K}}}\right)\textbf{V},
\end{equation}
where $d_{K}$ is the dimension of \textbf{K}. The MSA module uses the self-attention as multiple heads in parallel and concatenates the outputs followed by a linear projection as
\begin{equation}
    MSA(\textbf{Q}, \textbf{K}, \textbf{V}) = Concat(A_{1}, A_{2}, ..., A_{H})\textbf{W},
\end{equation}
where $H$ is the number of attention heads, $\textbf{W}$ $\in$ $\mathbb{R}^{H \times d_{k} \times N}$ is a learnable parameter matrix and $N$ is the number of patch tokens.
The output obtained by the MSA module is fed into the MLP module after the residual connection as $z = L(GELU(L(x)))$, 
where $x$ is the layer normalized output of the MSA and L($\cdot$) is a linear transformation operation. A dropout layer and layer normalization are applied after the MLP layer to reduce the gradient vanishing problem and to speed up training.

\subsection{3D-Convolution Guided Residual Module (3D-CGRM)}
Although ViTs excel at characterizing spectral signatures, they lack the encoding of spatial information since patches are handled as separate tokens. Therefore, we design a 3D-CGRM layer which uses a \texttt{3D-Conv} to fuse data across encoder layers and extract discriminative spectral features. 
The proposed CGRM is applied after every Transformer encoder.
Consider $\{e_{1},..., e_{L}\} \in \mathbb{R}^{N \times d_{t}}$ denote the output of the $L$ encoders. To implement the CGRM module, we unflatten the patches back into the original image and add a dimension to the tensor using the unsqueeze operation to produce $\{e'_{1},..., e'_{L-1}\} \in \mathbb{R}^{1 \times H \times W \times 64}$.  The tensors $e'_{l}$ and $e'_{l-1}$ are concatenated on the new dimension. The resultant tensor of dimension $\{2, H, W, 64\}$ is fed into a 3D-Convolution layer with kernel $(2, 3, 3)$ and padding $(0, 1, 1)$. This merges the respective spectral bands of the two encoder outputs and extracts the discriminative spectral features. The \texttt{3D-Conv} output $e''_{l} \in \mathbb{R}^{1 \times H \times W \times 64}$ is squeezed back to its original shape and flattened. The process of CGRM can be represented as
\begin{equation}
    \begin{aligned}
        e'_{l} &= Unsqueeze(Unflatten(e_{l})), \\
        e''_{l} &= Conv3d(concat[e'_{l-1},\hfill e'_{l}]), \\
        e_{l} & = Flatten(Squeeze(e''_{l})),
    \end{aligned}
\end{equation}
where $l = 1,..., L$. The output of CGRM is fed into the next encoder block. Layer normalization is applied to the output after the $L^{th}$ encoder block.

\subsection{Global Average Pool Classification}
In traditional ViT implementations, a learnable class (\texttt{CLS}) token is prefixed to the patch tokens and used to compute the final feature representation. However, we empirically find that the average pooled visual tokens contain more discriminative information than the single class token. Hence, we apply the Global Average Pooling on the final output visual token features as $c = L(Pool(e_{L}))$, where $e_{L}$ is the layer normalized output of the $L^{th}$ encoder layer and $L(\cdot)$ is a linear transformation. The shape after pooling is $(1 \times 64)$ which is transformed to $(1 \times C)$, where \texttt{C} is the number of classes.

\section{Experimental Settings}
\label{sec:experimental_settings}

\textbf{HSI datasets:} We use three benchmark HSI datasets, namely Houston, MUUFL and Botswana\footnote{The Houston and Botswana datasets can be found at: https://github.com/mhaut/HSI-datasets/} as summarized in Table \ref{dataset_stats}.
\textbf{\textit{Houston:}} The Houston 2013 HSI dataset is collected over the area surrounding the University of Houston. It consists of $15$ classes having $144$ spectral bands. It has dimensions of $340\times1905$ pixels with the spatial resolution being $2.5$ metres per pixel.
\textbf{\textit{MUUFL:}} The MUUFL dataset is collected by imaging the area over the University of Mississippi Gulfport. Its original dimensions are $325\times337$ pixels with $72$ spatial bands, with $11$ land cover classes. However, $8$ spectral bands are removed due to noise contamination, and dimensions are reduced to $320\times220$ pixels to account for the lost area. 
\textbf{\textit{Botswana:}} The Botswana dataset was acquired over the area surrounding the Okavango Delta in Botswana. The images contain $242$ spectral channels and dimensions of $1496\times256$ pixels. After removing the noisy and uncaliberated spectral bands we have $145$ spectral bands. To compensate for the lost area the spatial resolution is reduced to $1476\times256$. 

\begin{table}[!t]
\label{table:datasets}
\caption{The statistics of Houston, MUUFL and Botswana datasets using category and number of train and test samples.}
\resizebox{1\columnwidth}{!}{%
\begin{tabular}{p{0.022\textwidth}|p{0.09\textwidth}p{0.015\textwidth}p{0.027\textwidth}|p{0.06\textwidth}p{0.015\textwidth}p{0.028\textwidth}|p{0.093\textwidth}p{0.015\textwidth}p{0.02\textwidth}}
\hline
Class & Houston & Train & Test & MUUFL & Train & Test & Botswana & Train & Test    \\
\hline
1 & Healthy Grass & 198  & 1053 & Trees            & 1162  & 22084 & Water & 14  & 256  \\
2 & Stressed Grass  & 190  & 1064 & Mostly Grass     & 214   & 4056 & Hippo Grass           & 5   & 96  \\
3 & Synthetic Grass & 192  & 505 & Mixed Ground     & 344   & 6538 & Floodplains Grasses 1 & 13  & 238  \\
4 & Trees           & 188  & 1056 & Dirt and Sand    & 91    & 1735 & Floodplains Grasses 2 & 11  & 204   \\
5 & Soil            & 186  & 1056 & Roads            & 334   & 6353 & Reeds1                & 13  & 256   \\
6 & Water           & 182  & 143 & Water            & 23    & 443 & Riparian              & 13  & 256    \\
7 & Residential     & 196  & 1072 & Building Shadows & 112   & 2121 & Firescar2             & 13  & 246 \\
8 & Commercial      & 191  & 1053 & Buildings        & 312   & 5928 & Island Interior       & 10  & 193 \\
9 & Road            & 193  & 1059 & Sidewalks        & 69    & 1316 & Acacia Woodlands      & 16  & 298 \\
10 & Highway         & 191  & 1036  & Yellow Curbs     & 9     & 174 & Acacia Shrublands     & 12  & 236 \\
11 & Railway         & 181  & 1054  & Cloth Panels     & 14    & 255 & Acacia Grasslands     & 15  & 290  \\
12 & Parking Lot 1    & 192  & 1041 & & & & Short Moplane         & 9   & 172 \\
13 & Parking Lot 2    & 184  & 285 & & & & Mixed Moplane         & 13  & 255  \\
14 & Tennis Court & 181  & 247 & & & & Exposed Soils & 5 & 90 \\
15 & Running Track & 187 & 473 & & & & & & \\ \hline
Total  &  & 2832 & 12197 & & 2684 & 51003 & & 255 & 3086 \\ \hline 
\end{tabular}%
}
\label{dataset_stats}
\end{table}

\begin{table}
\caption{Accuracy (\%) comparision on Houston dataset.}
\label{table:houston}
\resizebox{1\columnwidth}{!}{%
\begin{tabular}{p{0.07\columnwidth}|p{0.06\columnwidth}p{0.06\columnwidth}p{0.06\columnwidth}p{0.06\columnwidth}p{0.06\columnwidth}p{0.06\columnwidth}p{0.06\columnwidth}p{0.06\columnwidth}p{0.1\columnwidth}}
\hline
Class  & SVM   & RF    & 1D-CNN & 2D-CNN          & 3D-CNN          & RNN   & ViT   & Morph-Former             & 3D-ConvSST            \\ \hline
1  & 80.44           & 78.73 & 80.25 & 66.48 & 82.24          & 78.06 & 79.20           & \textbf{82.53}& 82.52\\
2  & 82.14           & 72.74 & 78.29 & 81.77 & 81.02          & \textbf{89.00}& 74.15           & 84.77          & 84.96   \\
3  & \textbf{100.0} & 97.23 & 70.50 & 43.37 & 68.91          & 63.37 & 99.21           & 94.26          & 97.42           \\
4  & 74.81           & 76.89 & 91.95 & 82.10 & \textbf{99.34} & 54.26 & 92.80           & 96.69          & 98.57            \\
5  & 96.97           & 89.96 & 93.37 & 86.17 & 98.01          & 87.12 & 93.66           & 97.54          & \textbf{100.0}\\
6  & 95.10           & 88.11 & 77.62 & 76.92 & 81.12          & 89.51 & \textbf{100.0} & 95.80          & \textbf{100.0}\\
7  & 73.13           & 83.40 & 67.44 & 78.64 & 75.65          & 27.61 & 83.68           & 89.46          & \textbf{96.54}\\
8  & 56.79           & 41.60 & 68.19 & 55.84 & 57.55          & 1.71  & 67.71           & 70.09          & \textbf{91.26}\\
9  & 85.55           & 75.45 & 74.22 & 62.70 & 74.98          & 48.44 & 78.85           & \textbf{86.69}& 83.66   \\
10 & 65.25           & 38.90 & 52.61 & 46.81 & 52.41          & 10.33 & 51.93           & 67.08& \textbf{68.24}\\
11 & 84.63           & 58.16 & 72.96 & 48.96 & 63.38          & 37.48 & \textbf{90.70}& 80.27          & 88.99   \\
12 & 94.43           & 80.60 & 89.72 & 51.39 & 81.08          & 49.95 & 71.95           & 96.25          & \textbf{96.63}\\
13 & 77.19           & 66.32 & 79.65 & 78.60 & 80.70          & 34.39 & 82.11           & 91.93          & \textbf{92.98}\\
14 & 91.90           & 80.16 & 79.76 & 79.35 & 81.38          & 99.60 & \textbf{100.0} & 95.55          & \textbf{100.0}  \\
15 & 88.16           & 94.29 & 74.21 & 22.20 & 74.21          & 55.60 & 98.10           & 93.23          & \textbf{100.0}  \\ \hline
OA       & 81.00           & 72.16 & 76.67 & 64.21 & 76.45          & 50.55 & 80.91           & 86.35          & \textbf{90.37}   \\
AA       & 83.10           & 74.84 & 76.72 & 64.09 & 76.80          & 55.10 & 84.27           & 88.14          & \textbf{92.12}   \\
Kappa    & 79.46           & 69.93 & 74.76 & 61.17 & 74.46          & 46.63 & 79.28           & 85.23          & \textbf{89.54}  \\ \hline
\end{tabular}%
}
\label{Table_Houston_Results}
\end{table}

\begin{table}
\caption{Accuracy (\%) comparision on MUUFL dataset.}
\label{table:MUUFL}
\resizebox{1\columnwidth}{!}{%
\begin{tabular}{p{0.07\columnwidth}|p{0.06\columnwidth}p{0.06\columnwidth}p{0.06\columnwidth}p{0.06\columnwidth}p{0.06\columnwidth}p{0.06\columnwidth}p{0.06\columnwidth}p{0.06\columnwidth}p{0.1\columnwidth}}
\hline
Class  & SVM   & RF    & 1D-CNN & 2D-CNN          & 3D-CNN          & RNN   & ViT   & Morph-Former             & 3D-ConvSST            \\ \hline
1  & 93.62 & 93.34 & 97.07 & 97.12          & 97.61          & 96.98 & 97.86 & \textbf{97.83}& 97.71 \\
2  & 42.60 & 64.52 & 74.46 & 78.21          & 71.92          & 77.37 & 81.09 & 89.47          & \textbf{91.69}\\
3  & 65.31 & 74.37 & 80.30 & 85.50          & 85.87          & 85.15 & 84.06 & 91.85          & \textbf{92.12}\\
4  & 40.52 & 58.44 & 83.40 & 86.40          & 92.68          & 86.05 & 87.84 & 92.51& \textbf{93.94}\\
5  & 52.40 & 82.07 & 91.59 & 92.76          & \textbf{95.20} & 91.42 & 94.35 & 93.56          & 92.58         \\
6  & 0.00  & 62.08 & 83.07 & 91.65          & 94.58          & 69.75 & 94.58 & \textbf{97.74}& \textbf{97.74}\\
7  & 44.32 & 48.33 & 84.96 & 85.34          & 79.87          & 82.74 & 87.79 & 87.08          & \textbf{90.05}\\
8  & 64.93 & 81.60 & 92.34 & 93.29          & 88.38          & 92.16 & 95.61 & 96.64          & \textbf{97.52}\\
9  & 0.00  & 14.51 & 52.43 & \textbf{63.98} & 53.72          & 55.24 & 62.01 & 61.63          & 62.23          \\
10 & 0.00  & 3.45  & 18.39 & \textbf{30.46} & 22.41          & 20.69 & 21.84 & 22.99          & 13.21        \\
11 & 61.57 & 70.20 & 53.73 & 45.49          & 45.49          & 57.65 & 78.82 & \textbf{80.39}& 77.64 \\ \hline
OA       & 69.90 & 80.06 & 89.16 & 90.89          & 90.11          & 89.92 & 91.99 & 93.82          & \textbf{94.10}\\
AA       & 69.90 & 59.35 & 73.79 & 77.29          & 75.25          & 74.11 & 80.53 & \textbf{82.88} & 82.40          \\
Kappa    & 58.96 & 73.29 & 85.58 & 87.92          & 86.83          & 86.62 & 89.38 & 91.82          & \textbf{92.20}\\ \hline
\end{tabular}%
}
\label{Table_MUUFL_Results}
\end{table}

\begin{table}[!t]
\caption{Accuracy (\%) comparision on Botswana dataset.}
\label{table:botswana}
\resizebox{1\columnwidth}{!}{%
\begin{tabular}{p{0.07\columnwidth}|p{0.06\columnwidth}p{0.06\columnwidth}p{0.06\columnwidth}p{0.06\columnwidth}p{0.06\columnwidth}p{0.06\columnwidth}p{0.06\columnwidth}p{0.06\columnwidth}p{0.1\columnwidth}}
\hline
Class  & SVM   & RF    & 1D-CNN & 2D-CNN          & 3D-CNN          & RNN   & ViT   & Morph-Former             & 3D-ConvSST            \\ \hline
1  & 99.61           & 96.09 & 98.83           & \textbf{100.0} & 94.14 & \textbf{100.0} & 97.66     & \textbf{100.0} & 99.21\\
2  & \textbf{100.0} & 93.75 & 80.21           & 86.46           & 40.63 & 87.50           & \textbf{100.0} & \textbf{100.0} & 96.87\\
3  & 98.32           & 94.12 & \textbf{100.0} & \textbf{100.0} & 89.08 & 76.05           & \textbf{100.0} & \textbf{100.0} & \textbf{100.0} \\
4  & 97.55           & 88.73 & 99.51           & \textbf{100.0} & 83.82 & 84.31           & \textbf{100.0} & \textbf{100.0} & \textbf{100.0} \\
5  & 85.94           & 58.98 & 80.86           & 62.11           & 70.70 & 85.55           & 91.80           & 95.31& \textbf{96.09}\\
6  & 87.89           & 68.75 & 80.47           & 89.84           & 51.17 & 81.64           & 92.58           & \textbf{96.48}  & 95.70\\
7  & \textbf{100.0} & 99.59 & \textbf{100.0} & \textbf{100.0} & 95.53 & 99.59           & \textbf{100.0} & \textbf{100.0} & \textbf{100.0} \\
8  & \textbf{100.0} & 93.26 & 98.45           & 98.96           & 69.95 & 78.24           & \textbf{100.0} & \textbf{100.0} & \textbf{100.0} \\
9  & 95.30           & 76.17 & 96.31           & 77.85           & 94.97 & 87.92           & 97.32           & 96.98           & \textbf{99.66}  \\
10 & 99.15           & 83.05 & 92.80           & 91.95           & 64.83 & 69.07           & \textbf{100.0} & 99.58           & \textbf{100.0} \\
11 & \textbf{100.0} & 98.97 & 85.52           & 89.31           & 83.45 & 95.52           & \textbf{100.0} & \textbf{100.0} & \textbf{100.0} \\
12 & 93.60           & 91.28 & 91.28           & 92.44           & 84.88 & 95.35           & 98.84           & 98.84           & \textbf{100.0} \\
13 & 93.33           & 85.88 & 88.24           & 92.16           & 37.65 & 82.75           & 97.65           & \textbf{100.0} & \textbf{100.0}\\
14 & \textbf{82.22}& \textbf{82.22}& 81.11           & 85.56           & 0.00  & 78.89           & \textbf{82.22}& \textbf{82.22}& \textbf{82.22}\\ \hline
OA       & 95.56           & 85.97 & 91.67           & 90.28           & 73.40 & 86.36           & 97.47           & 98.41           & \textbf{98.60}\\
AA       & 95.21           & 86.49 & 90.97           & 90.47           & 68.63 & 85.88           & 97.00           & 97.82           & \textbf{97.84}\\
Kappa    & 95.19           & 84.80 & 90.98           & 89.47           & 71.09 & 85.21           & 97.26           & 98.28           & \textbf{98.49}\\ \hline
\end{tabular}%
}
\label{Table_Botswana_Results}
\end{table}

\textbf{Evaluation Metrics:} Overall accuracy (OA), average accuracy (AA), kappa coefficient, and categorical classification accuracy  are used to objectively analyse the classification performance. Furthermore, the classification maps obtained by different models are visualised for qualitative comparison.

\textbf{Experimental Setup:} The validation tests are carried out using a Google Colab environment which has an Intel Xeon CPU with 13 GB of RAM and a Tesla K80 accelerator. The batch size is set to $64$. The training is performed for $500$ epochs with Adam optimizer and initial learning rate $5e^{-4}$. However, the 3D-ConvSST can converge with fewer epochs. The input image size for the 3D-ConvSST is set to $11\times11$ and the patch size is set to $1\times1$. The Transformer encoder receives the tokens with $64$ dimension. The MLP has the same number of heads and size as in Morphformer \cite{morphformer}. Each encoder contains an MLP with two fully connected layers as well as a GELU activation function after the first layer. Following the MLPs, a dropout layer drops $10\%$ of the neurons.

\begin{figure*}[!t]
\label{fig:houston}
    \centering
    \begin{subfigure}[b]{0.329\textwidth}
    \includegraphics[width=\textwidth,height=1.1cm]{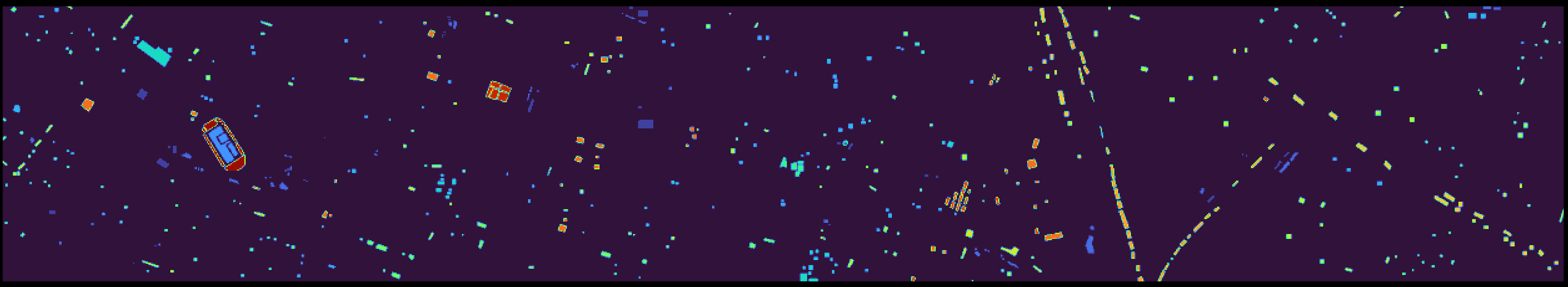}
    \caption{Ground Truth}
    \end{subfigure}
    \begin{subfigure}[b]{0.665\textwidth}  \includegraphics[width=\textwidth,height=1.1cm]{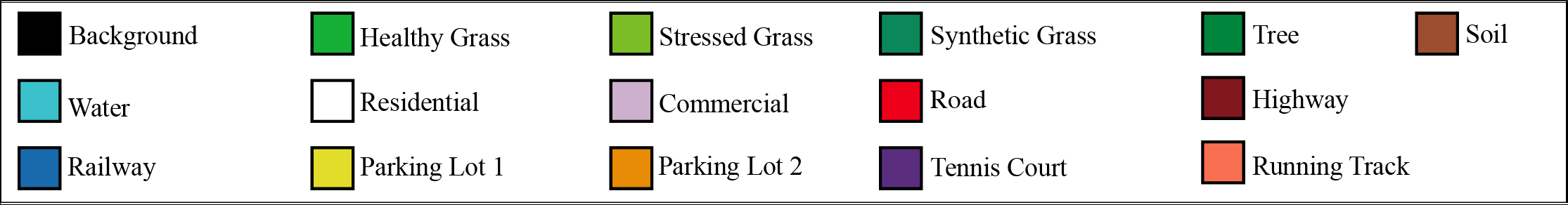}
    \caption{Class Labels}
    \end{subfigure}
    \begin{subfigure}[b]{0.329\textwidth}
    \includegraphics[width=\textwidth,height=1.1cm]{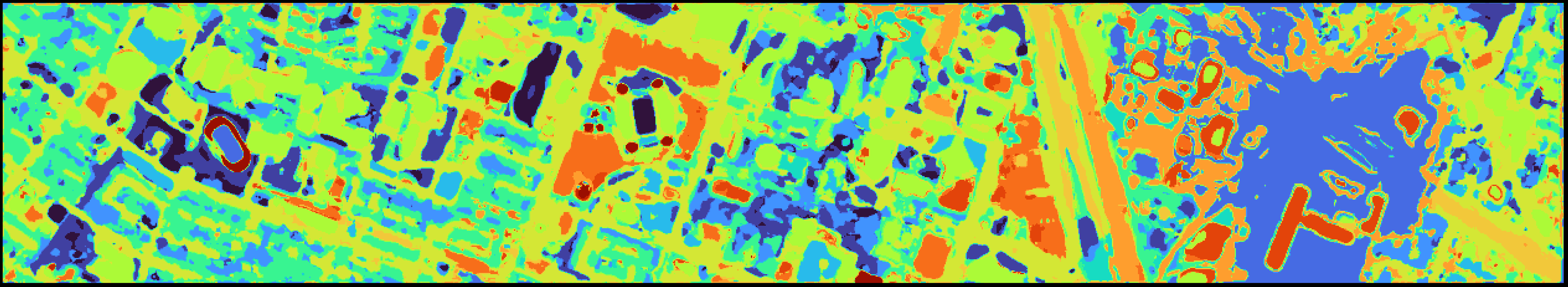}
    \caption{SVM}
    \end{subfigure}
    \begin{subfigure}[b]{0.329\textwidth}
    \includegraphics[width=\textwidth,height=1.1cm]{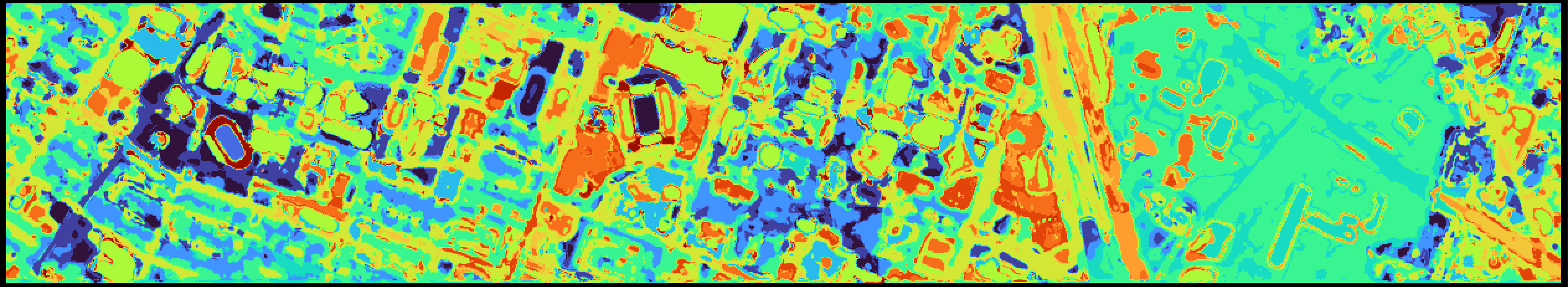}
    \caption{Random Forest (RF)}
    \end{subfigure}
    \begin{subfigure}[b]{0.329\textwidth}
    \includegraphics[width=\textwidth,height=1.1cm]{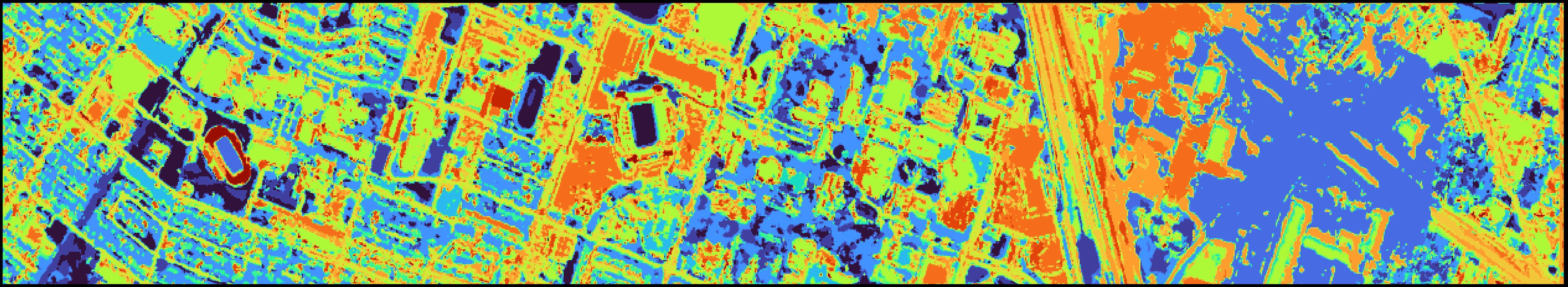}
    \caption{1D-CNN}
    \end{subfigure}
    \begin{subfigure}[b]{0.329\textwidth}
    \includegraphics[width=\textwidth,height=1.1cm]{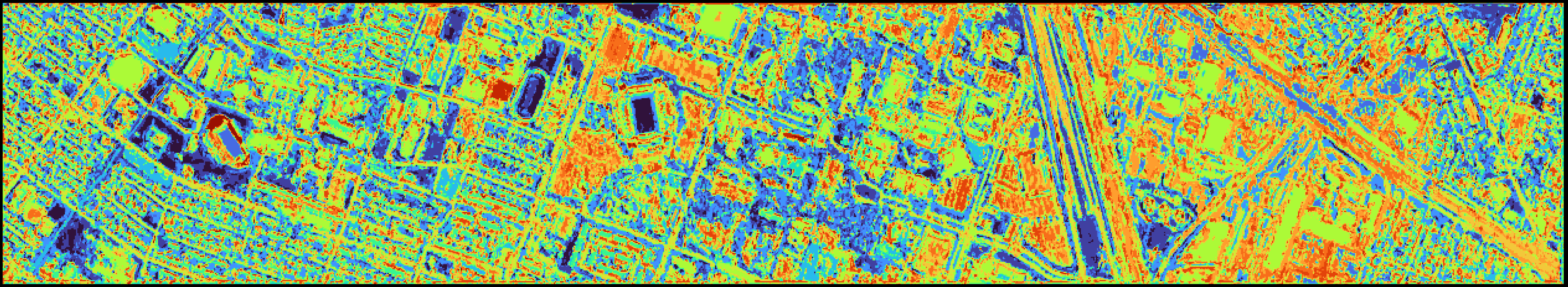}
    \caption{2D-CNN}
    \end{subfigure}
    \begin{subfigure}[b]{0.329\textwidth}
    \includegraphics[width=\textwidth,height=1.1cm]{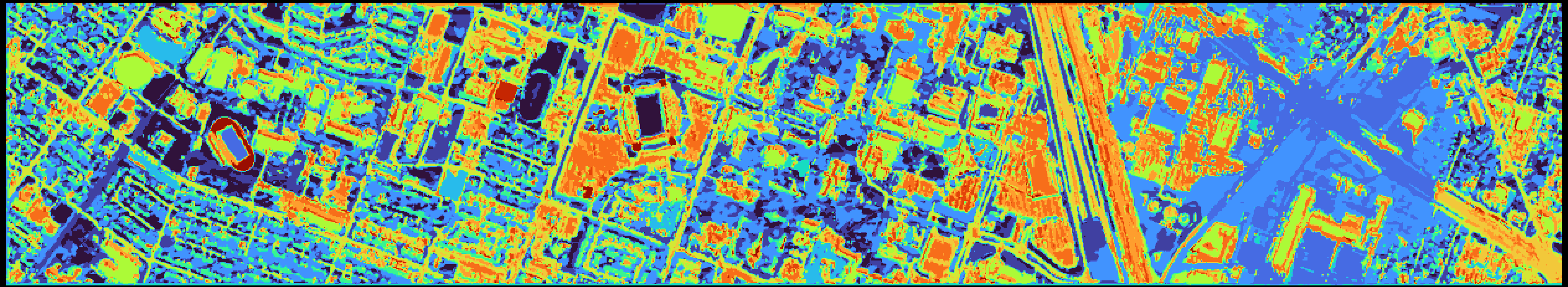}
    \caption{3D-CNN}
    \end{subfigure}
    \begin{subfigure}[b]{0.329\textwidth}
    \includegraphics[width=\textwidth,height=1.1cm]{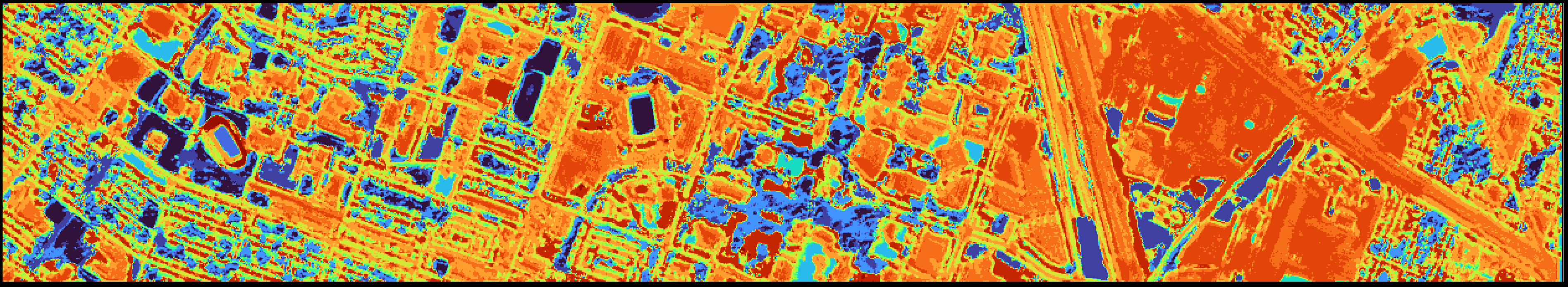}
    \caption{RNN}
    \end{subfigure}
    \begin{subfigure}[b]{0.329\textwidth}
    \includegraphics[width=\textwidth,height=1.1cm]{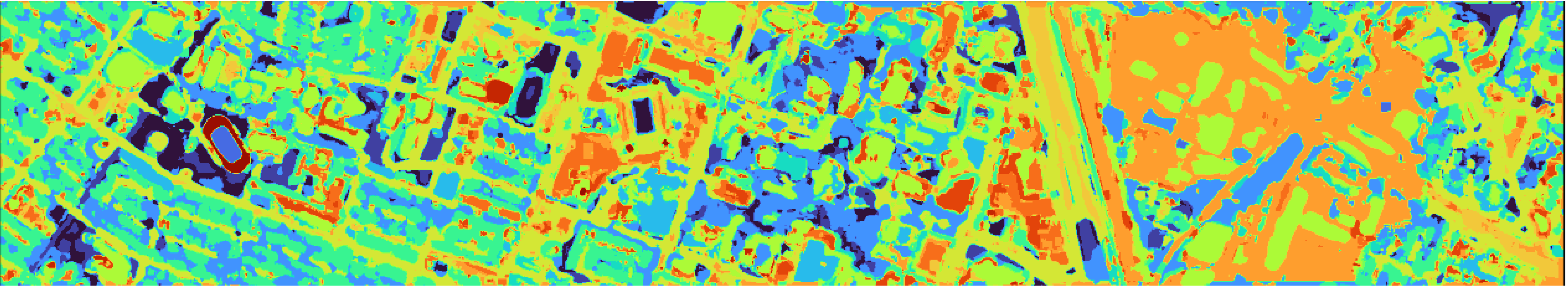}
    \caption{ViT}
    \end{subfigure}
    \begin{subfigure}[b]{0.329\textwidth}
    \includegraphics[width=\textwidth,height=1.1cm]{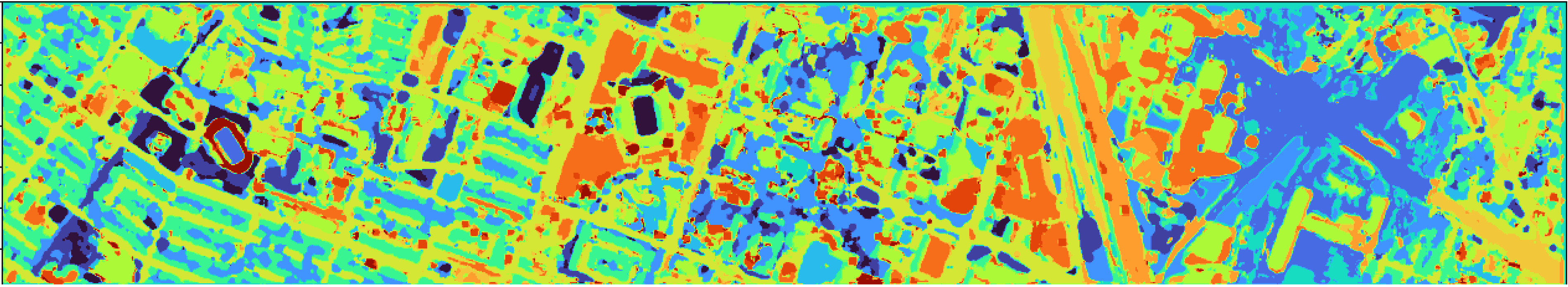}
    \caption{MorphFormer}
    \end{subfigure}
    \begin{subfigure}[b]{0.329\textwidth}
    \includegraphics[width=\textwidth,height=1.1cm]{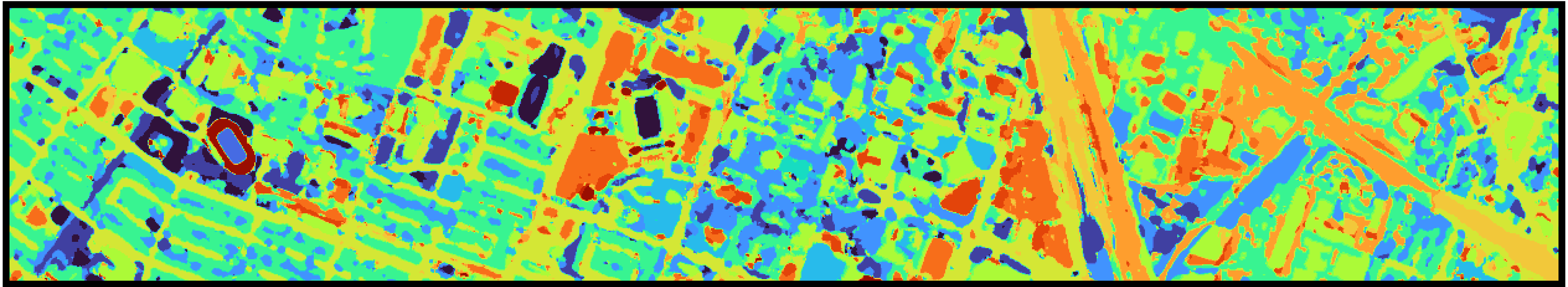}
    \caption{3D-ConvSST}
    \end{subfigure}
    \caption{Visualization maps for the University of Houston (UH) HSI Dataset}
    \label{fig:CMap_Houston}
\end{figure*}

\textbf{Compared Methods:} Extensive tests are carried out using the proposed 3D-ConvSST model and the results are compared to those of traditional and state-of-the-art models. The base code used for the compared methods is from \cite{ahmad2021hyperspectral} GitHub repository\footnote{The source code for the comparitive methods: https://github.com/AnkurDeria/HSI-Traditional-to-Deep-Models}.  
Traditional classifiers such as Support Vector Machine (SVM) \cite{svm} and Random Forest (RF) \cite{rf} are compared. 
The standard deep learning models are also compared, such as 1D-CNN \cite{cnn1d}, 2D-CNN \cite{cnn2d}, 3D-CNN \cite{cnn3d}, and RNN \cite{rnn}. 
We also compare with cutting-edge Transformer-based approaches, such as Vision Transformer (ViT) \cite{visionTransformer} and MorphFormer \cite{morphformer}. The ViT model uses $8$ transformer encoder blocks and an initial learning rate of $4.95e^{-4}$.

\begin{figure*}[!t]
\label{fig:MUUFL}
    \centering
    \begin{subfigure}[b]{0.161\textwidth}
\includegraphics[width=\textwidth,height=2cm]{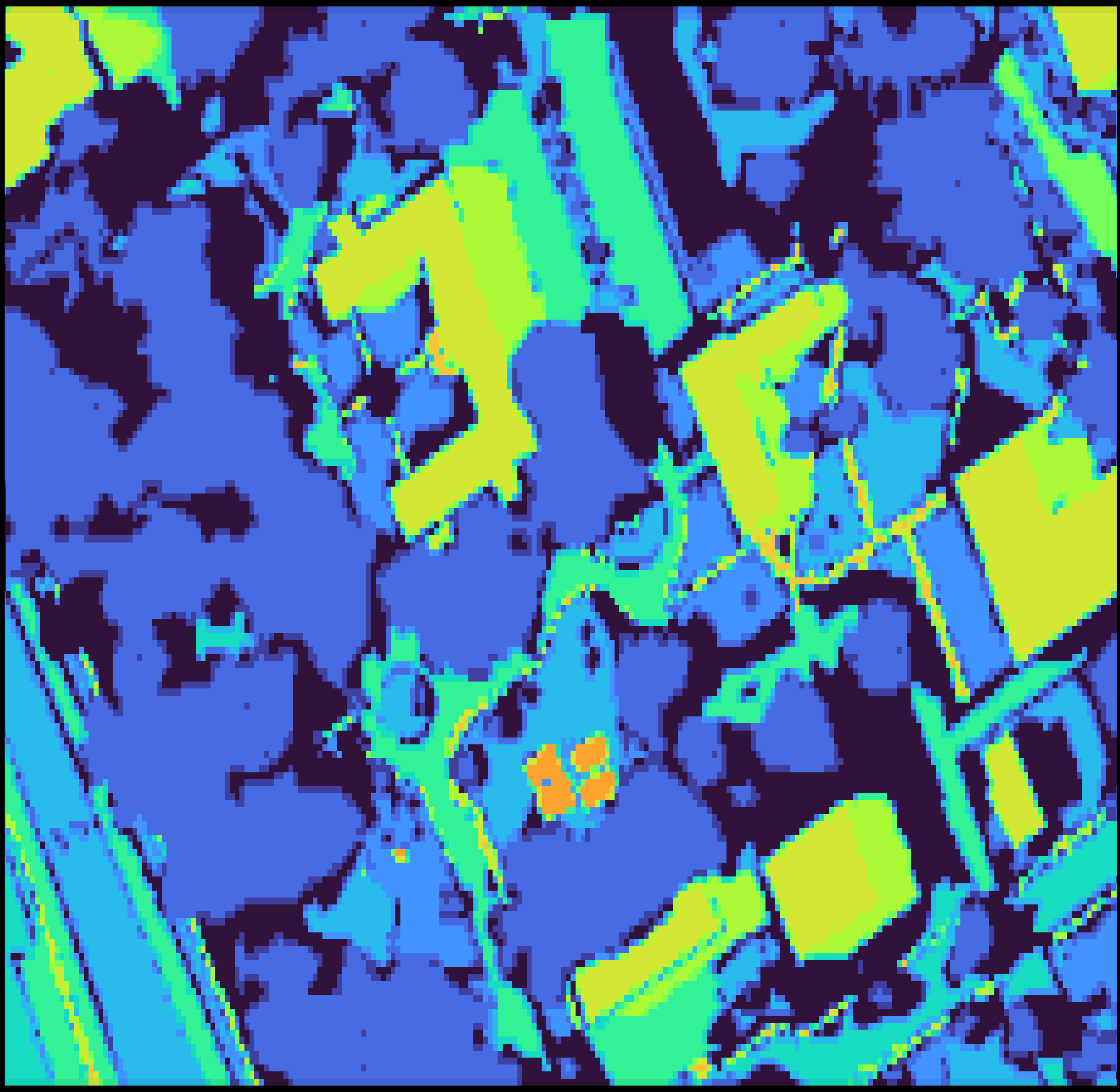}
    \caption{Ground Truth}
    \end{subfigure}
    \begin{subfigure}[b]{0.161\textwidth}
    \includegraphics[width=\textwidth,height=2cm]{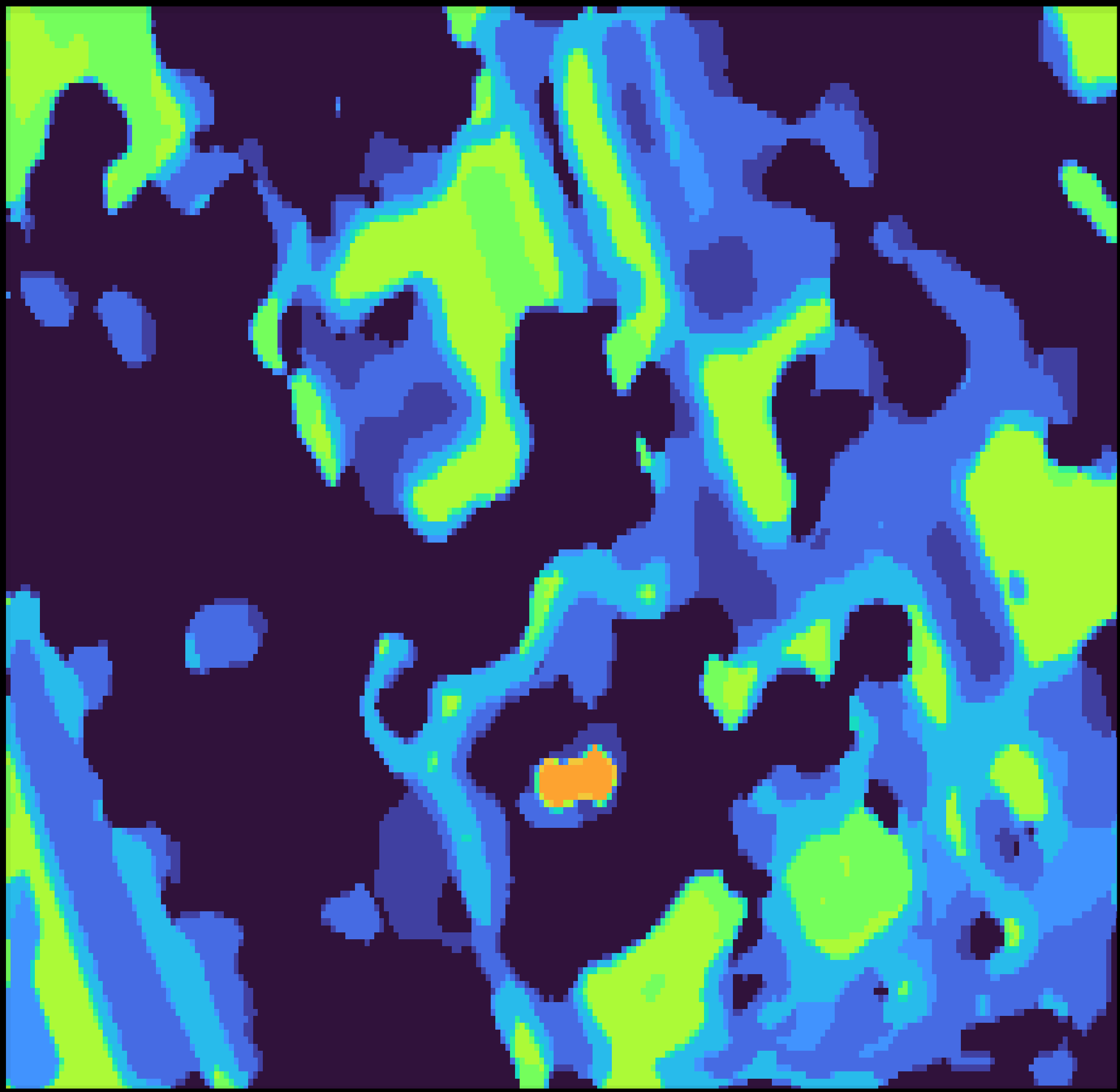}
    \caption{SVM}
    \end{subfigure}
    \begin{subfigure}[b]{0.161\textwidth}
    \includegraphics[width=\textwidth,height=2cm]{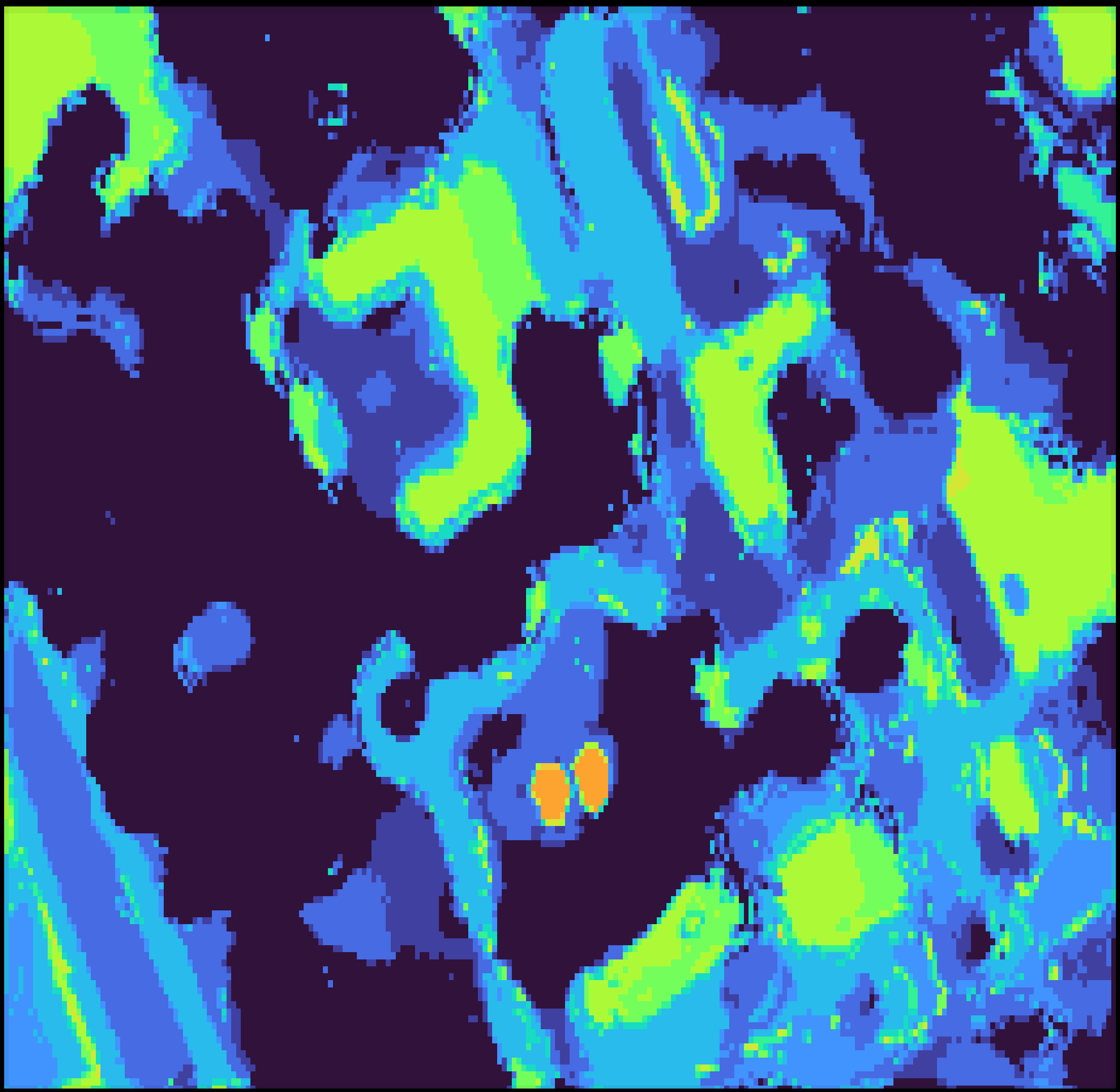}
    \caption{RF}
    \end{subfigure}
    \begin{subfigure}[b]{0.161\textwidth}
    \includegraphics[width=\textwidth,height=2cm]{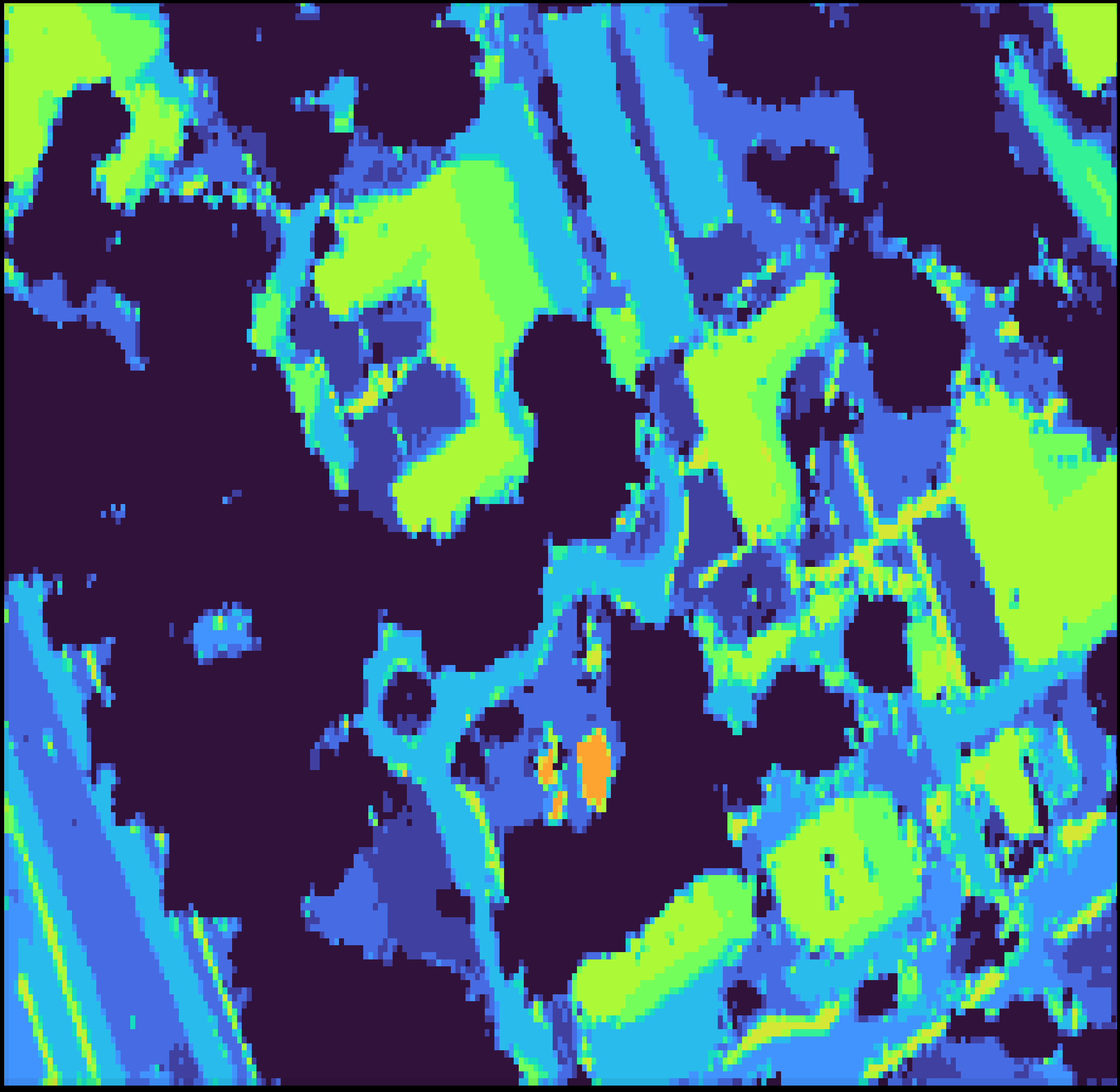}
    \caption{1D-CNN}
    \end{subfigure}
    \begin{subfigure}[b]{0.161\textwidth}
    \includegraphics[width=\textwidth,height=2cm]{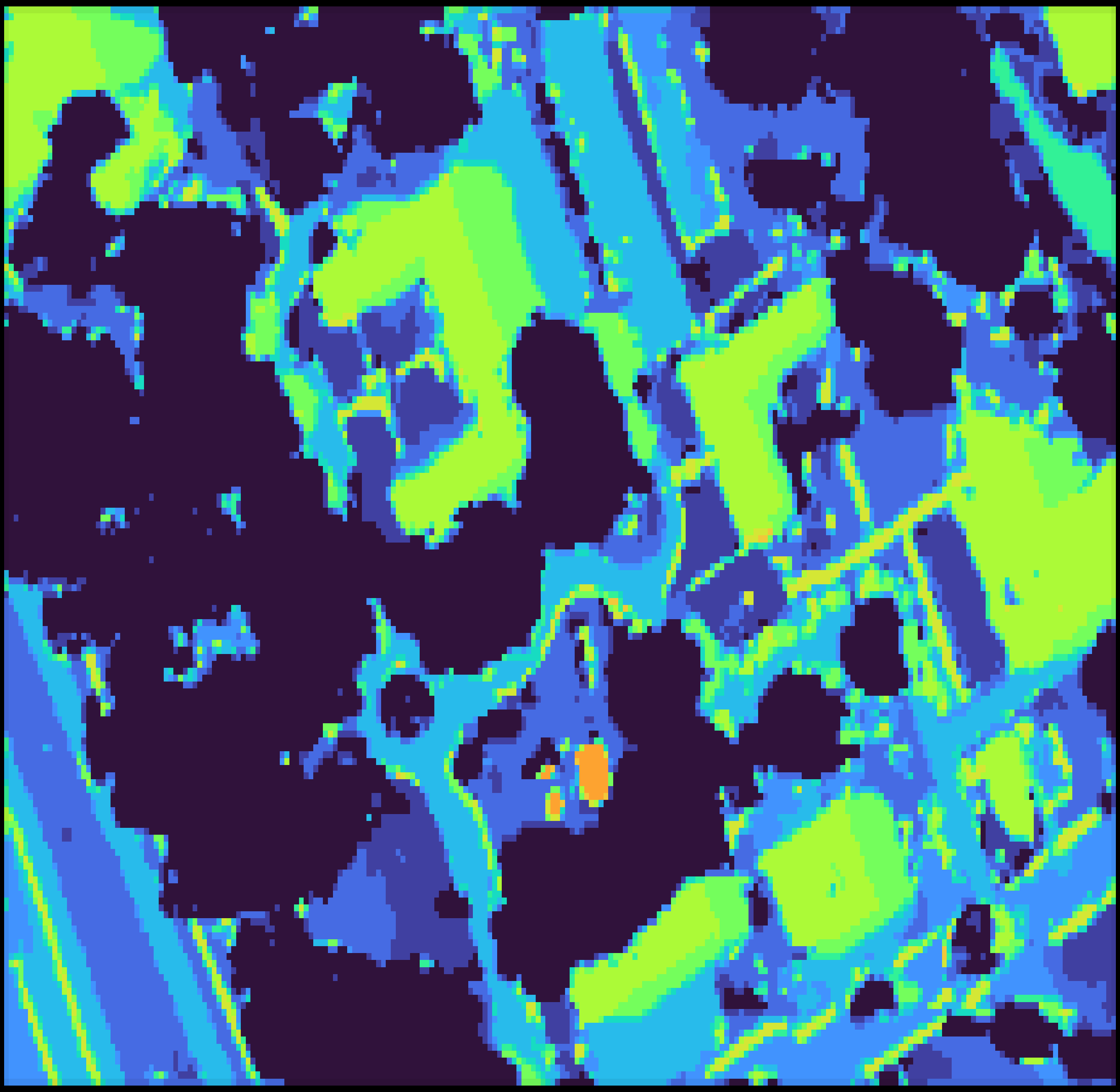}
    \caption{2D-CNN}
    \end{subfigure}
    \begin{subfigure}[b]{0.161\textwidth}
    \includegraphics[width=\textwidth,height=2cm]{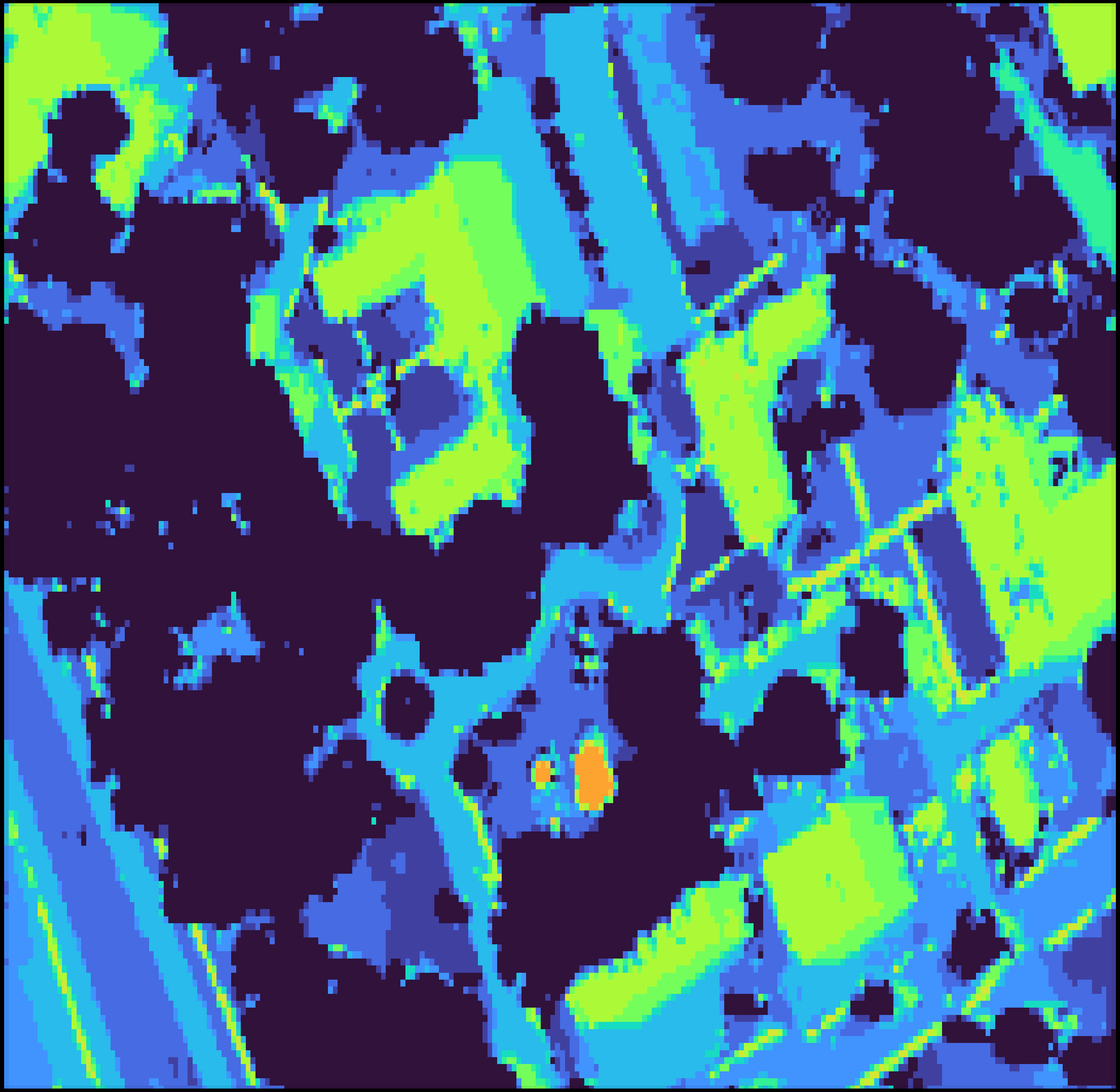}
    \caption{3D-CNN}
    \end{subfigure}
    \begin{subfigure}[b]{0.161\textwidth}
    \includegraphics[width=\textwidth,height=2cm]{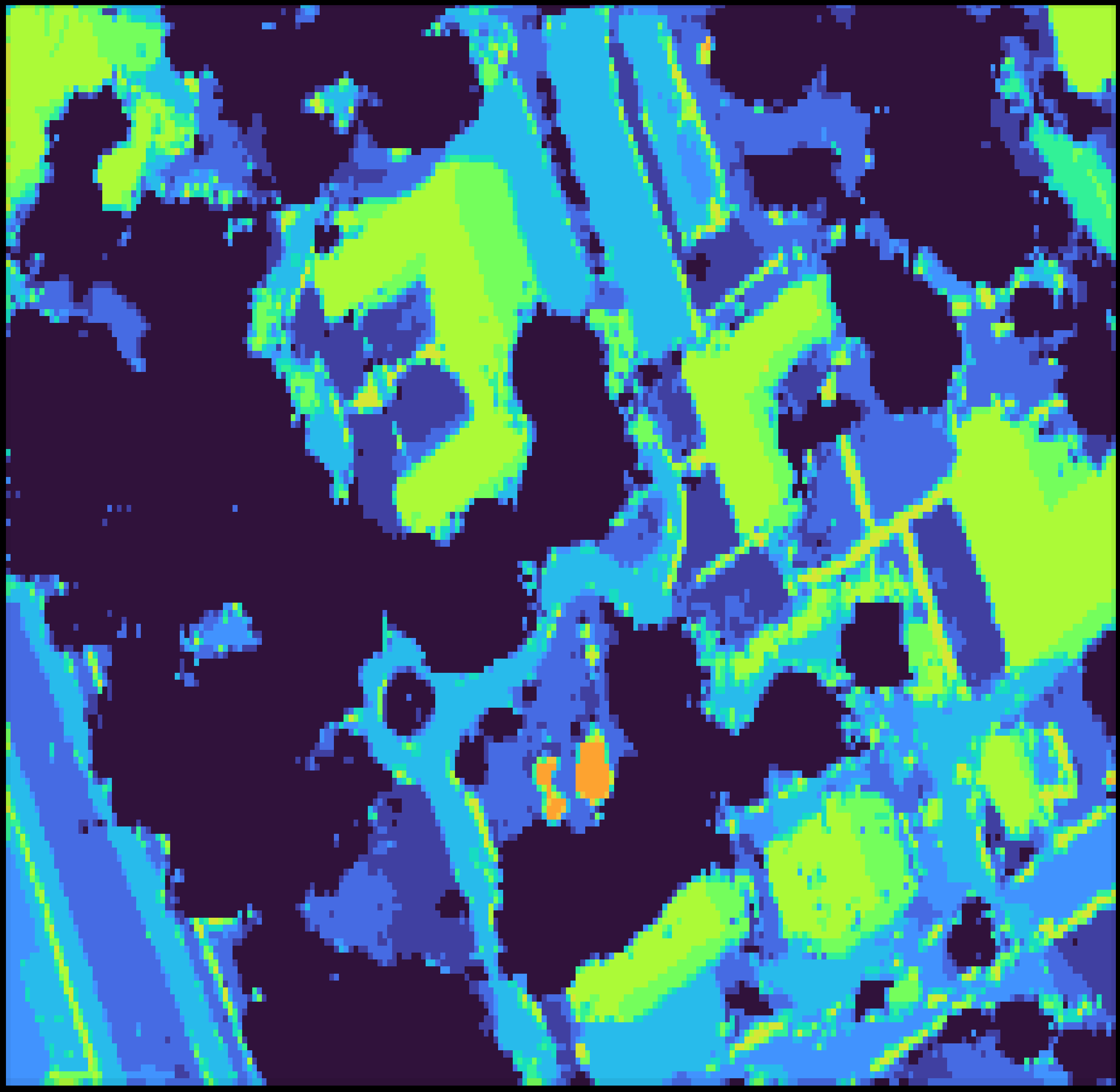}
    \caption{RNN}
    \end{subfigure}
    \begin{subfigure}[b]{0.161\textwidth}
    \includegraphics[width=\textwidth,height=2cm]{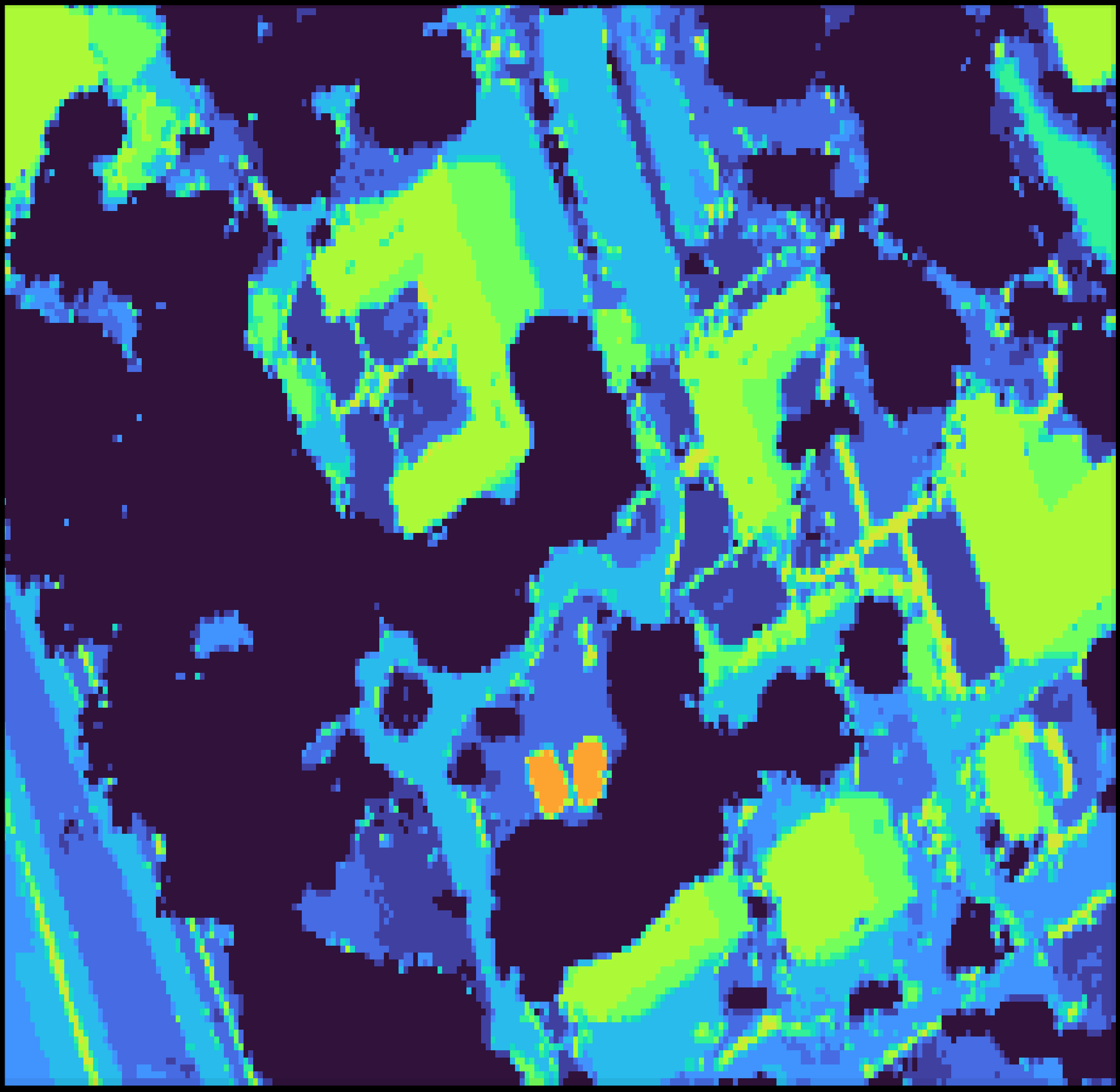}
    \caption{ViT}
    \end{subfigure}
    \begin{subfigure}[b]{0.161\textwidth}
    \includegraphics[width=\textwidth,height=2cm]{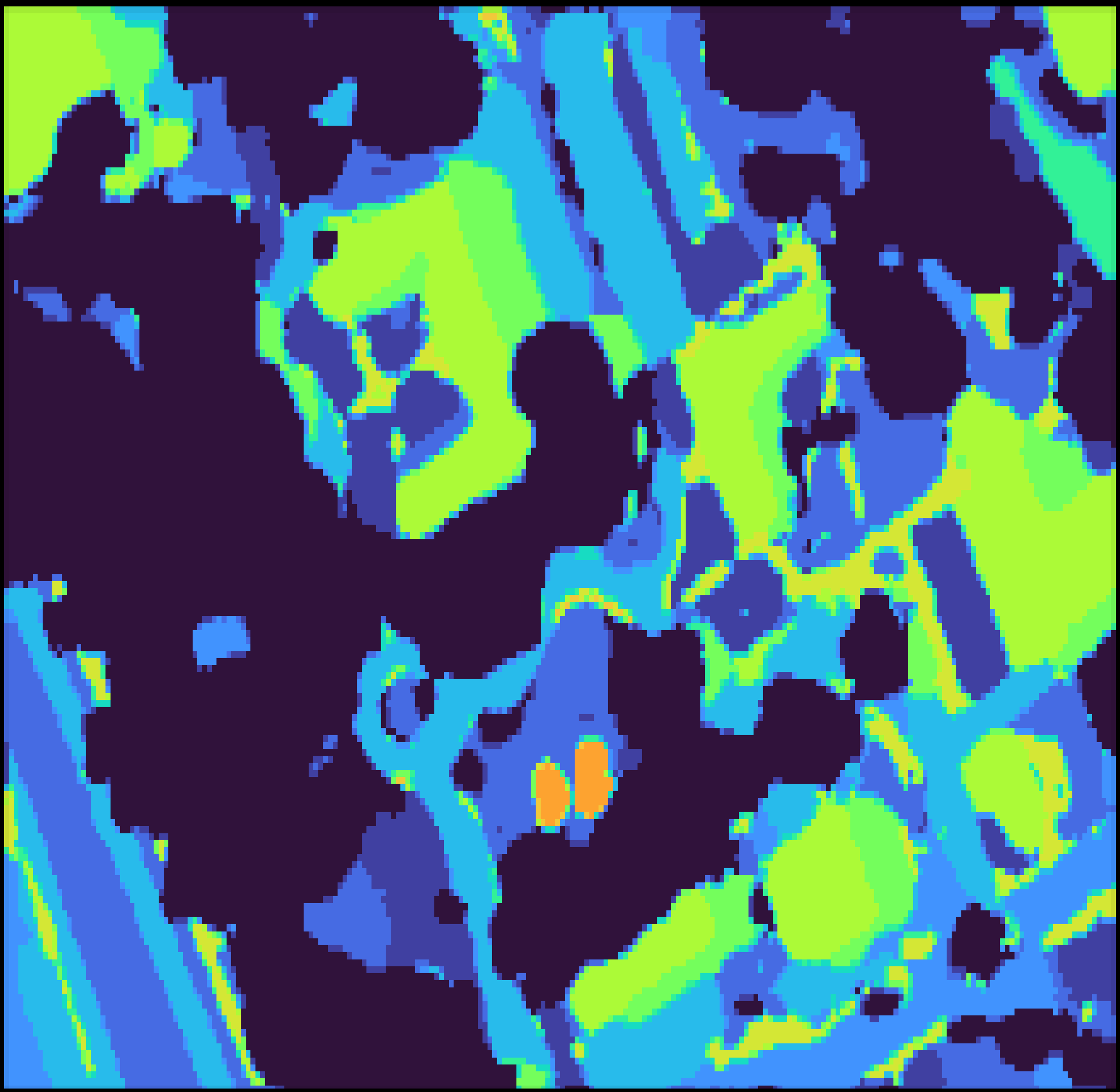}
    \caption{MorphFormer}
    \end{subfigure}
    \begin{subfigure}[b]{0.161\textwidth}
    \includegraphics[width=\textwidth,height=2cm]{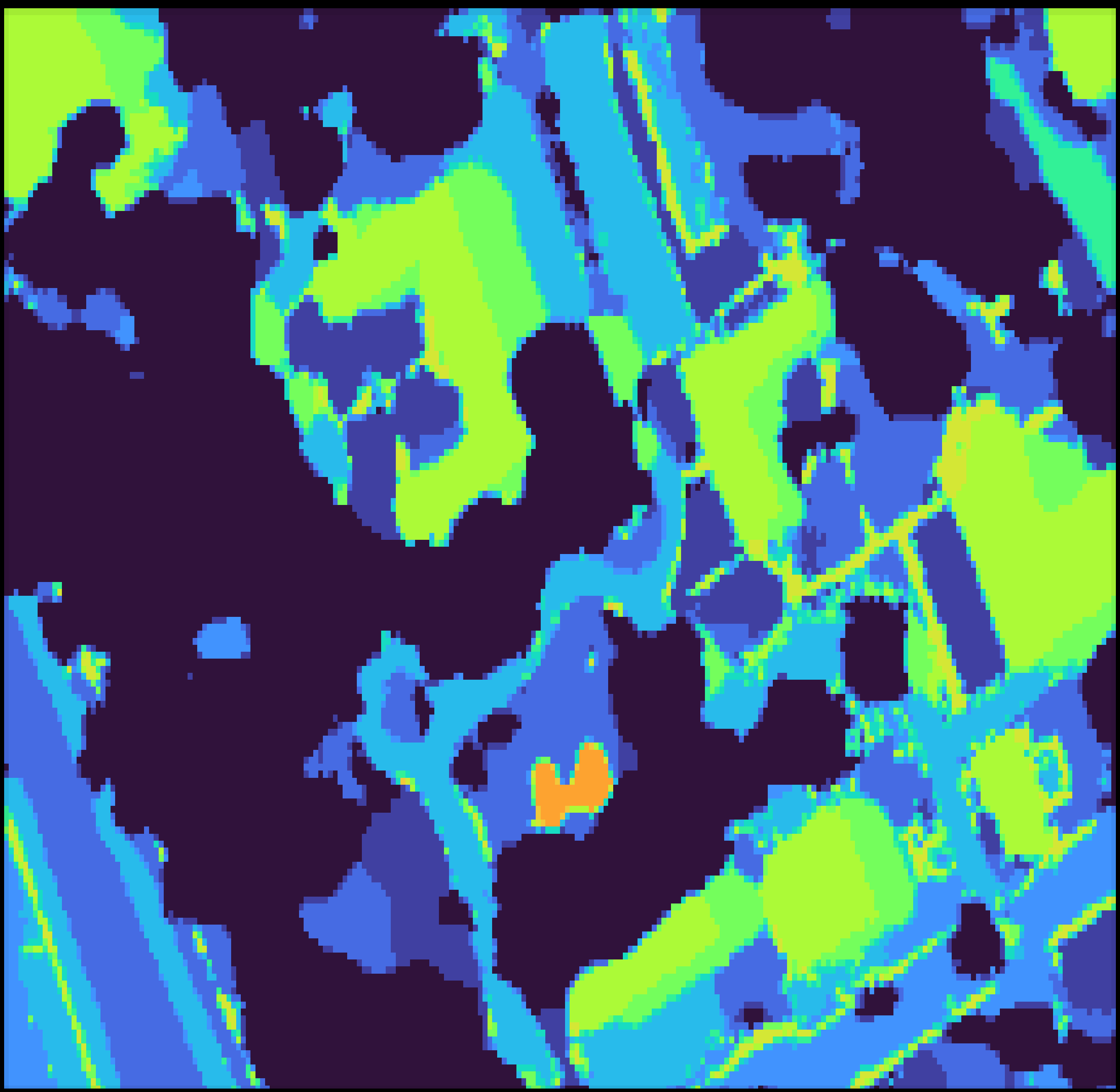}
    \caption{3D-ConvSST}
    \end{subfigure}
    \begin{subfigure}[b]{0.322\textwidth}  \includegraphics[width=\textwidth,height=2cm,trim={0cm 0cm 0cm 0cm},clip]{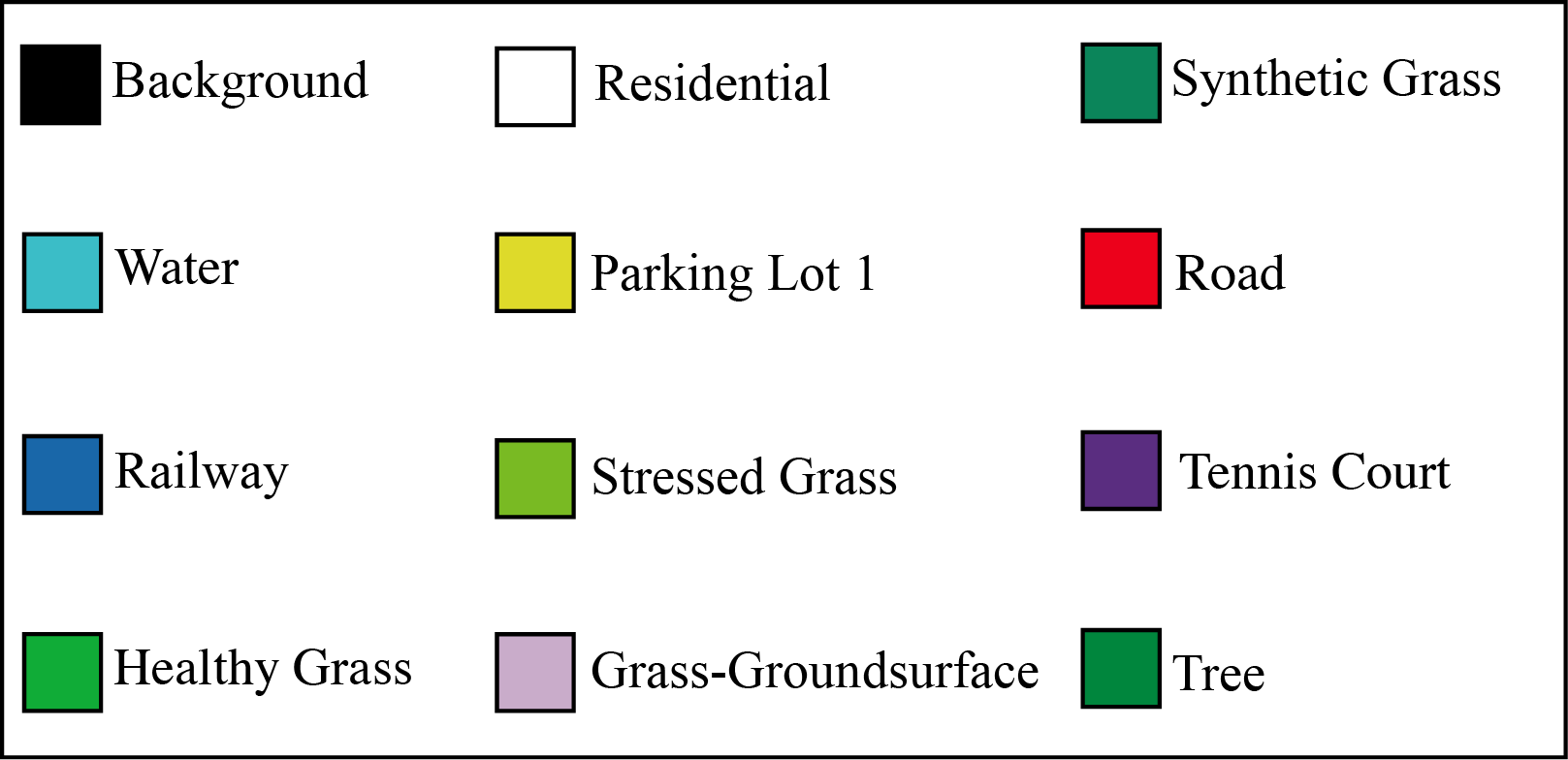}
    \caption{Class Labels}
    \end{subfigure}
    \caption{Visualization maps for the MUUFL HSI Dataset}
    \label{fig:CMap_MUUFL}
\end{figure*}

\begin{figure*}[!t]
\label{fig:botswana}
    \centering
    \begin{subfigure}[b]{0.329\textwidth}
    \includegraphics[width=\textwidth,height=1.2cm]{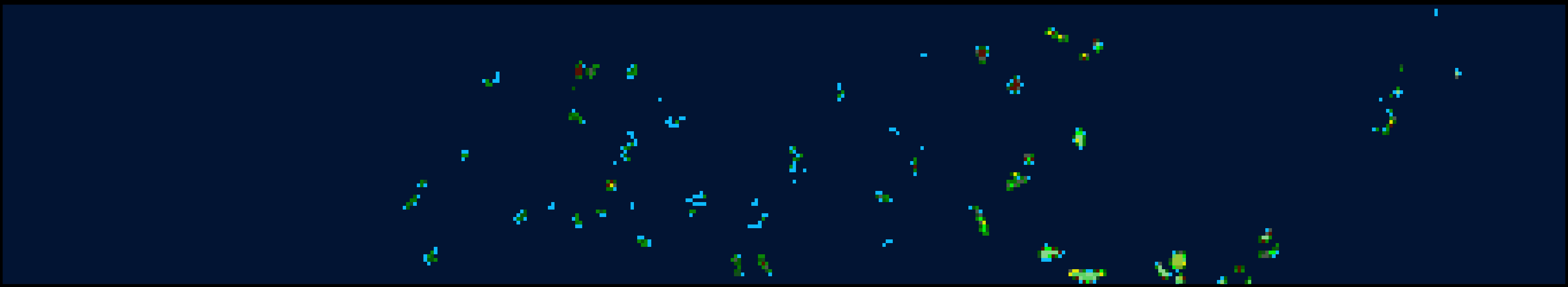}
    \caption{Ground Truth}
    \end{subfigure}
    \begin{subfigure}[b]{0.665\textwidth}  \includegraphics[width=\textwidth,height=1.2cm]{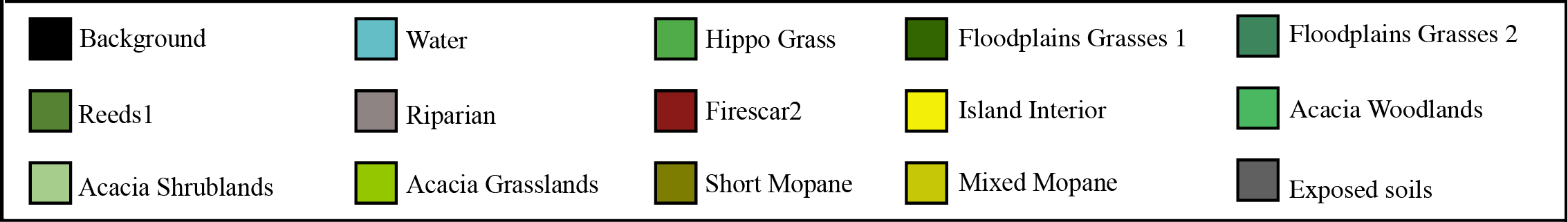}
    \caption{Class Labels}
    \end{subfigure}
    \begin{subfigure}[b]{0.329\textwidth}
    \includegraphics[width=\textwidth,height=1.2cm]{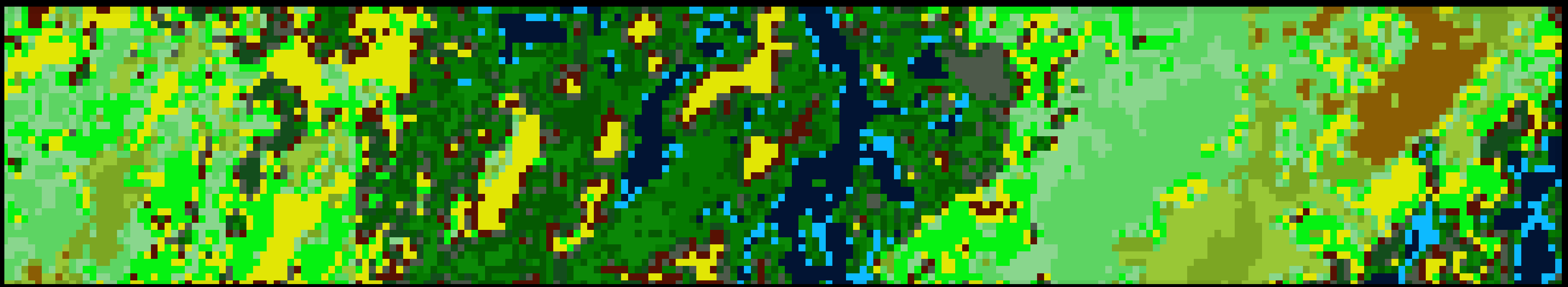}
    \caption{SVM}
    \end{subfigure}
    \begin{subfigure}[b]{0.329\textwidth}
    \includegraphics[width=\textwidth,height=1.2cm]{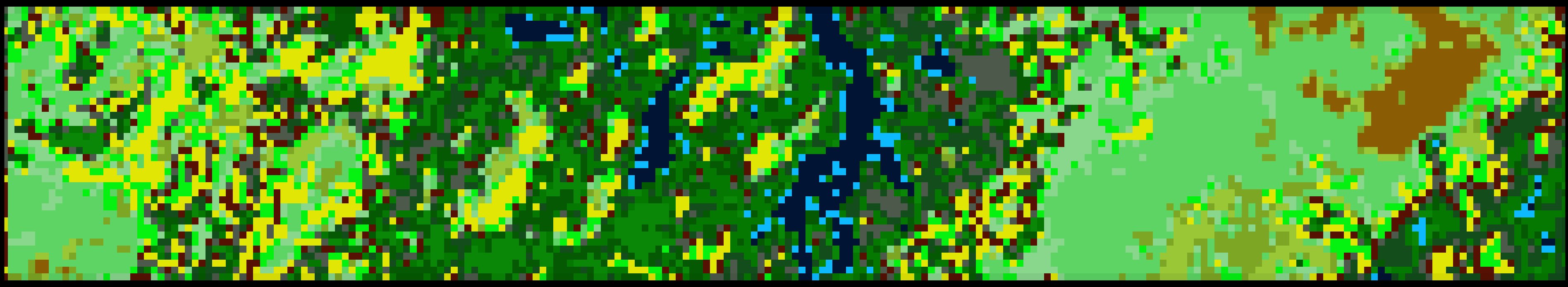}
    \caption{Random Forest (RF)}
    \end{subfigure}
    \begin{subfigure}[b]{0.329\textwidth}
    \includegraphics[width=\textwidth,height=1.2cm]{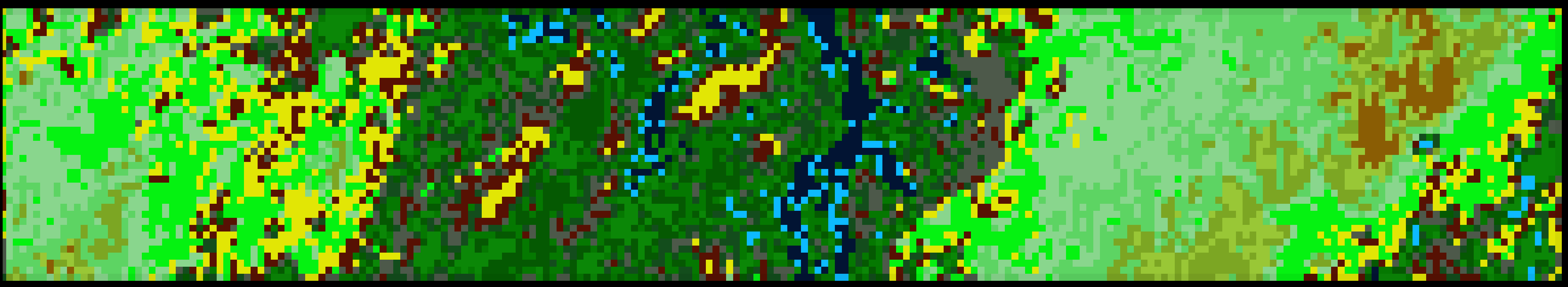}
    \caption{1D-CNN}
    \end{subfigure}
    \begin{subfigure}[b]{0.329\textwidth}
    \includegraphics[width=\textwidth,height=1.2cm]{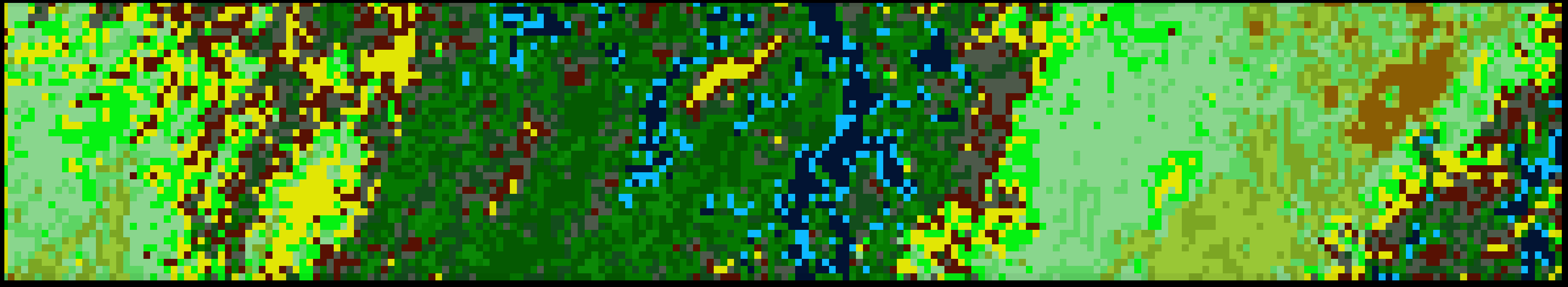}
    \caption{2D-CNN}
    \end{subfigure}
    \begin{subfigure}[b]{0.329\textwidth}
    \includegraphics[width=\textwidth,height=1.2cm]{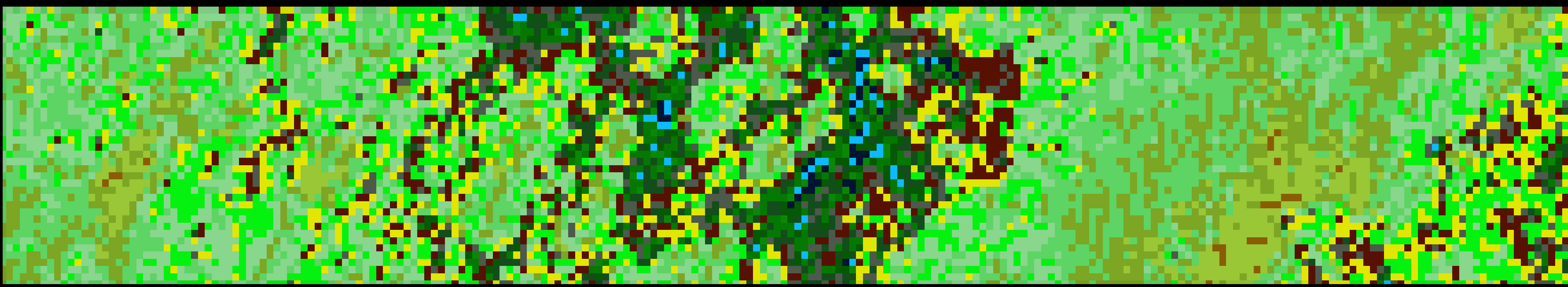}
    \caption{3D-CNN}
    \end{subfigure}
    \begin{subfigure}[b]{0.329\textwidth}
    \includegraphics[width=\textwidth,height=1.2cm]{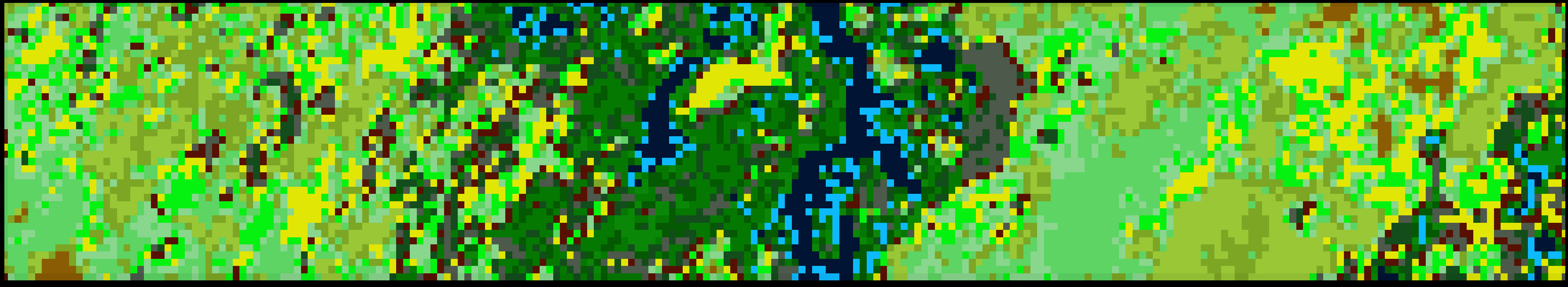}
    \caption{RNN}
    \end{subfigure}
    \begin{subfigure}[b]{0.329\textwidth}
    \includegraphics[width=\textwidth,height=1.2cm]{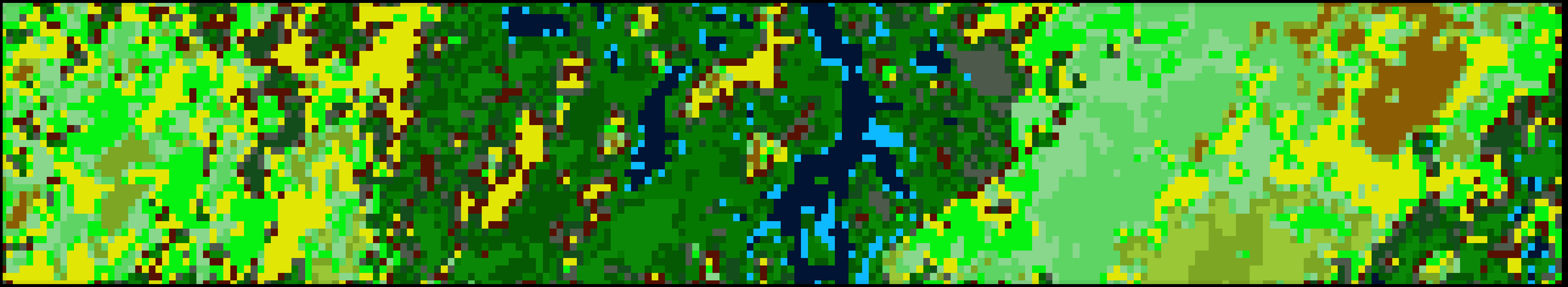}
    \caption{ViT}
    \end{subfigure}
    \begin{subfigure}[b]{0.329\textwidth}
    \includegraphics[width=\textwidth,height=1.2cm]{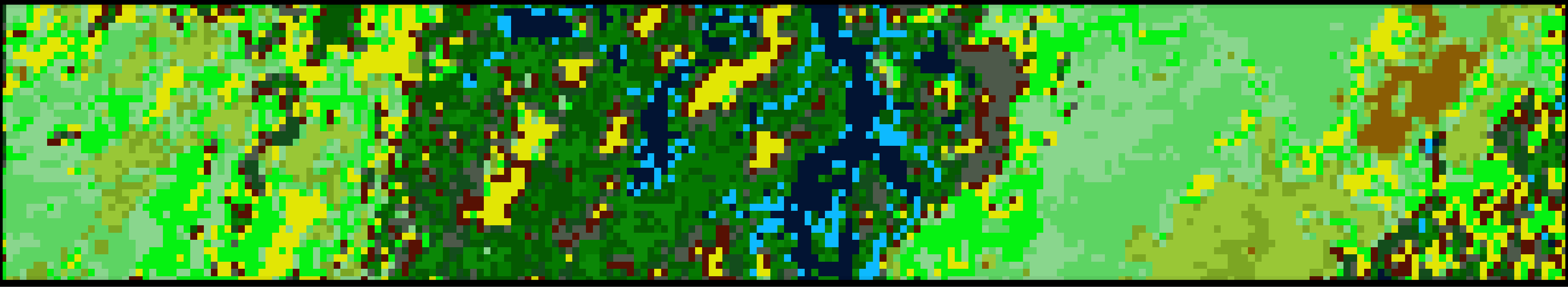}
    \caption{MorphFormer}
    \end{subfigure}
    \begin{subfigure}[b]{0.329\textwidth}
    \includegraphics[width=\textwidth,height=1.2cm]{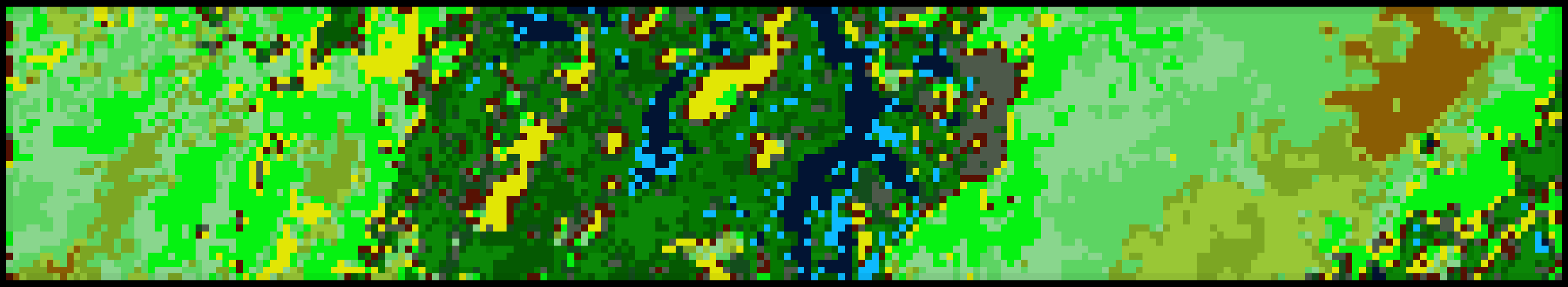}
    \caption{3D-ConvSST}
    \end{subfigure}
    \caption{Visualization maps for the Botswana HSI Dataset}
    \label{fig:CMap_Botswana}
\end{figure*}

\section{Experimental Results and Analysis}
\label{sec:experimental}

\subsection{Experimental Quantitative Results}
Tables \ref{Table_Houston_Results}, \ref{Table_MUUFL_Results}, and \ref{Table_Botswana_Results} present the quantitative evaluation of classification performance. The best classification results are highlighted in bold. The results show that our proposed method outperforms all other techniques in terms of OA, AA, and Kappa while being superior to other methods when comparing class-wise accuracy. It is worth mentioning that CNN-based classifiers perform similarly, with the exception of 2D-CNN on the Houston dataset and 3D-CNN on the Botswana dataset which show notably worse metrics. The SVM classifier performs well on the Houston and Botswana datasets compared to the Random Forest classifier, but fell short on the MUUFL dataset, misclassifying three classes. RNNs performance is on par with the Convolution-based classifiers, but unable to converge on the Houston dataset. Due to the attention mechanism, Transformer approaches such as ViT and MorphFormer function better on all three datasets. However, incorporating spatial-spectral information into the proposed 3D-ConvSST improves classification performance in terms of OA, AA, and Kappa on all the datasets.
Table \ref{Table_Houston_Results} demonstrates that the SVM could outperform the conventional and deep learning-based classifiers on the Houston dataset and on par with traditional ViT. However, the MorphFormer and 3D-ConvSST models perform better than other models. Because of its improved ability to learn spatial and spectral information, the proposed method outperforms all other models. 
Table \ref{Table_MUUFL_Results} displays the MUUFL dataset's generalisation ability for fragmented train and test samples. All of the deep learning techniques outperform the traditional classifiers in terms of accuracy. The 3D-ConvSST outperforms every model, including Transformer-based approaches. 
Table \ref{Table_Botswana_Results} lists the classification results on the Botswana dataset. SVM outperforms all the conventional and deep learning classifiers. However, the results of Transformer models are better with the outstanding performance of the proposed 3D-ConvSST model.

\begin{figure}[!t]
\label{fig:ablation}
\centering
\includegraphics[width=.33\columnwidth]{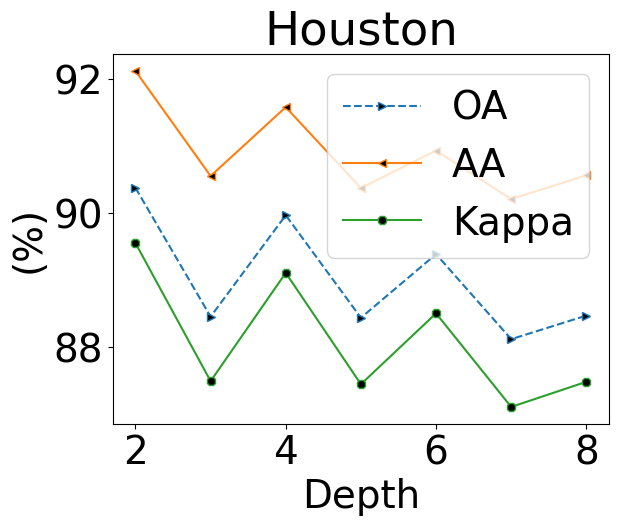}\hfill
\includegraphics[width=.33\columnwidth]{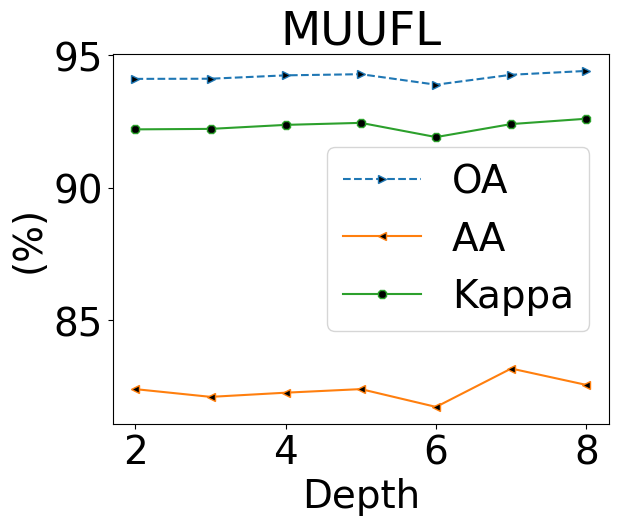}\hfill
\includegraphics[width=.33\columnwidth]{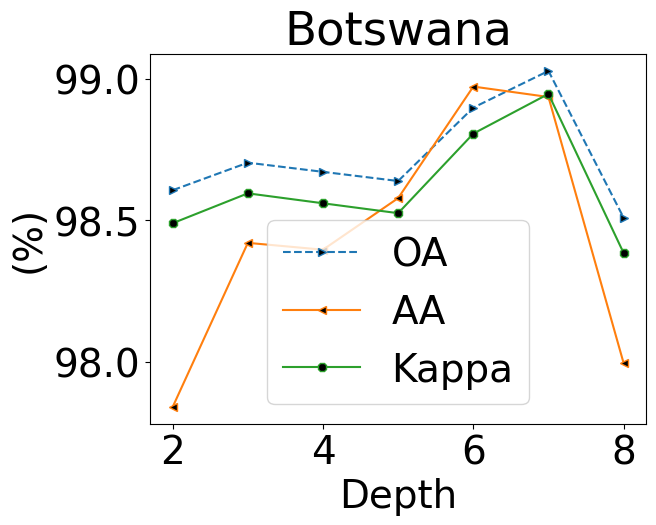}
\caption{Effect of the 3D-ConvSST encoder depth. 
}
\label{fig:encoder_depth}
\end{figure}

\subsection{Visual Comparison}
Figs. \ref{fig:CMap_Houston}, \ref{fig:CMap_MUUFL} and \ref{fig:CMap_Botswana} show the obtained classification maps on Houston, MUUFL, and Botswana datasets respectively. Conventional classifiers like SVM and RF only utilize spectral information and generally provide classification maps with salt and pepper noise around the edges. The same is also noticeable in the CNN-based classifiers. This is indicative of the inability of the classifiers to accurately identify the objects. The Transformer models provide smoother classification maps, as it can extract more abstract information with sequential representation. Compared to ViT and MorphFormer, our 3D-ConvSST model provides the best classification maps due to its improved spatial-spectral fusion method to characterize the texture and edge details.

\subsection{Performance over Different Encoder Depths}
Fig. \ref{fig:encoder_depth} shows the classification performance of the 3D-ConvSST model over different encoder depths on Houston, MUUFL and Botswana datasets. There is a tradeoff since deeper Transformer models increase the complexity of the model, but susceptible to overfitting. 
In the Houston dataset, a depth of two encoders achieves the best OA, AA and Kappa values. MUUFL and Botswana prefer deeper models, achieving the highest classification performance at encoder depths of $8$ and $7$, respectively. However, on the MUUFL dataset, the difference in classification performance across each depth is very minuscule compared to that in Houston and Botswana.

\begin{table}[!t]
\caption{Ablation analysis on the 3D-ConvSST model.}
\label{table:ablation}
\centering
\resizebox{0.95\columnwidth}{!}{%
\begin{tabular}{c|cc|ccc}
\hline
Dataset & \multicolumn{2}{c|}{Modules} & \multicolumn{3}{c}{Metrics} \\\cline{2-6}
& CGRM & AvgPool & OA (\%) & AA (\%) & Kappa (\%) \\\hline
\multirow{4}{*}{ Houston } & \xmark & \xmark & 87.86 & 89.21 & 86.82 \\
& \xmark & \cmark & 87.30 & 89.61 & 86.22 \\
& \cmark & \xmark & 87.82 & 89.63 & 86.80 \\
& \cmark & \cmark & \textbf{90.37} & \textbf{92.12} & \textbf{89.55} \\\hline
\multirow{4}{*}{ MUUFL } & \xmark & \xmark & 93.89 & 81.06 & 91.90 \\
& \xmark & \cmark & 93.27 & 79.39 & 91.07 \\
& \cmark & \xmark & 93.59 & 79.31 & 91.50 \\
& \cmark & \cmark & \textbf{94.11} & \textbf{82.41} & \textbf{92.20} \\\hline
\multirow{4}{*}{ Botswana } & \xmark & \xmark & 98.54 & 97.71 & 98.42 \\
& \xmark & \cmark & \textbf{98.61} & \textbf{98.03} & \textbf{98.49} \\
& \cmark & \xmark & 98.51 & 97.98 & 98.38 \\
& \cmark & \cmark & \textbf{98.61} & 97.84 & \textbf{98.49} \\\hline
\end{tabular}%
}
\label{Table_Ablation}
\end{table}

\subsection{Ablation Analysis}
Classification performance of the 3D-ConvSST is analysed by progressively adding CGRM and AvgPool modules to the architecture on the Houston, MUUFL and Botswana datasets as showcased in Table \ref{Table_Ablation}.
The model showcases similar performance when using one or neither of the two modules. In case of using only the CGRM module, the extracted spatial characteristics of the Conv3D layer cannot be fully utilized by the CLS token. In case of using only the AvgPool, the Average Pooling layer cannot showcase its full potential without the discriminative spatial characteristics extracted by the CGRM module. However, using both the modules together showcases a significant increase in accuracy in Houston and MUUFL. On the Botswana dataset, the performance using both is equal to using only AvgPool as this dataset contains more discriminative spectral features compared to spatial features. Overall, both modules together provide a significant improvement (up to $9\%$ OA) compared to a traditional ViT.

\section{Conclusion}
\label{sec:conclusion}
In this paper, we present a novel 3D-ConvSST architecture based on improved spatial-spectral fusion within the Transformer model with 3D-Convolution guided residual module (CGRM) and global average pooling (AvgPool). The CGRM encodes the spatial-spectral features of subsequent Transformer encoders. However, the AvgPool captures the spatial discriminative context in the final feature representation.
The use of both CGRM and AvgPool exploits the complimentary information and showcases superior HSI classification performance compared to other deep learning and conventional classifiers. The higher encoder depth is preferred on MUUFL and Botswana datasets and lower on Houston dataset.

{\small
\bibliographystyle{IEEEtran}
\bibliography{References}

\begin{thebibliography}{10}
\providecommand{\url}[1]{#1}
\csname url@samestyle\endcsname
\providecommand{\newblock}{\relax}
\providecommand{\bibinfo}[2]{#2}
\providecommand{\BIBentrySTDinterwordspacing}{\spaceskip=0pt\relax}
\providecommand{\BIBentryALTinterwordstretchfactor}{4}
\providecommand{\BIBentryALTinterwordspacing}{\spaceskip=\fontdimen2\font plus
\BIBentryALTinterwordstretchfactor\fontdimen3\font minus
  \fontdimen4\font\relax}
\providecommand{\BIBforeignlanguage}[2]{{%
\expandafter\ifx\csname l@#1\endcsname\relax
\typeout{** WARNING: IEEEtran.bst: No hyphenation pattern has been}%
\typeout{** loaded for the language `#1'. Using the pattern for}%
\typeout{** the default language instead.}%
\else
\language=\csname l@#1\endcsname
\fi
#2}}
\providecommand{\BIBdecl}{\relax}
\BIBdecl

\bibitem{nonconvexmodel}
D.~Hong, W.~He, N.~Yokoya, J.~Yao, L.~Gao, L.~Zhang, J.~Chanussot, and X.~Zhu,
  ``Interpretable hyperspectral artificial intelligence: When nonconvex
  modeling meets hyperspectral remote sensing,'' \emph{IEEE Geoscience and
  Remote Sensing Magazine}, vol.~9, no.~2, pp. 52--87, 2021.

\bibitem{frontiersspatialspectral}
P.~Ghamisi, E.~Maggiori, S.~Li, R.~Souza, Y.~Tarablaka, G.~Moser, A.~De~Giorgi,
  L.~Fang, Y.~Chen, M.~Chi, S.~Serpico, and J.~Benediktsson, ``New frontiers in
  spectral-spatial hyperspectral image classification: The latest advances
  based on mathematical morphology, markov random fields, segmentation, sparse
  representation, and deep learning,'' \emph{IEEE Geoscience and Remote Sensing
  Magazine}, vol.~6, pp. 10--43, 09 2018.

\bibitem{hsiurban}
J.~Yuan, S.~Wang, C.~Wu, and Y.~Xu, ``Fine-grained classification of urban
  functional zones and landscape pattern analysis using hyperspectral satellite
  imagery: A case study of wuhan,'' \emph{IEEE Journal of Selected Topics in
  Applied Earth Observations and Remote Sensing}, vol.~15, pp. 3972--3991,
  2022.

\bibitem{hsiagri}
B.~Lu, Y.~He, and P.~D. Dao, ``Comparing the performance of multispectral and
  hyperspectral images for estimating vegetation properties,'' \emph{IEEE
  Journal of Selected Topics in Applied Earth Observations and Remote Sensing},
  vol.~12, no.~6, pp. 1784--1797, 2019.

\bibitem{hsienvironment}
S.~Ustin, \emph{Manual Of Remote Sensing / Remote Sensing For Natural Resource
  Management And Environmental Monitoring}, 01 2004.

\bibitem{hsisvm}
F.~Melgani and L.~Bruzzone, ``Classification of hyperspectral remote sensing
  images with support vector machines,'' \emph{IEEE Transactions on Geoscience
  and Remote Sensing}, vol.~42, no.~8, pp. 1778--1790, 2004.

\bibitem{randomforesthsi}
J.~Ham, Y.~Chen, M.~Crawford, and J.~Ghosh, ``Investigation of the random
  forest framework for classification of hyperspectral data,'' \emph{IEEE
  Transactions on Geoscience and Remote Sensing}, vol.~43, no.~3, pp. 492--501,
  2005.

\bibitem{hsioverview}
S.~Li, W.~Song, L.~Fang, Y.~Chen, P.~Ghamisi, and J.~A. Benediktsson, ``Deep
  learning for hyperspectral image classification: An overview,'' \emph{IEEE
  Transactions on Geoscience and Remote Sensing}, vol.~57, no.~9, pp.
  6690--6709, 2019.

\bibitem{overviewDL}
------, ``Deep learning for hyperspectral image classification: An overview,''
  \emph{IEEE Transactions on Geoscience and Remote Sensing}, vol.~57, no.~9,
  pp. 6690--6709, 2019.

\bibitem{cnn}
K.~Makantasis, K.~Karantzalos, A.~Doulamis, and N.~Doulamis, ``Deep supervised
  learning for hyperspectral data classification through convolutional neural
  networks,'' in \emph{2015 IEEE International Geoscience and Remote Sensing
  Symposium (IGARSS)}, 2015, pp. 4959--4962.

\bibitem{cnn1d}
W.~Hu, Y.~Huang, L.~Wei, F.~Zhang, and H.~Li, ``Deep convolutional neural
  networks for hyperspectral image classification,'' \emph{Journal of Sensors},
  vol. 2015, pp. 1--12, 2015.

\bibitem{cnn2d}
W.~Zhao and S.~Du, ``Spectral–spatial feature extraction for hyper- spectral
  image classificationspectral–spatial feature extraction for hyper- spectral
  image classification,'' \emph{IEEE Transactions on Geoscience and Remote
  Sensing}, vol.~55, p. 4729–4742, 2017.

\bibitem{branch}
J.~Yang, Y.-Q. Zhao, and J.~C.-W. Chan, ``Learning and transferring deep joint
  spectral--spatial features for hyperspectral classification,'' \emph{IEEE
  Transactions on Geoscience and Remote Sensing}, vol.~55, no.~8, pp.
  4729--4742, 2017.

\bibitem{branch1}
X.~Xu, W.~Li, Q.~Ran, Q.~Du, L.~Gao, and B.~Zhang, ``Multisource remote sensing
  data classification based on convolutional neural network,'' \emph{IEEE
  Transactions on Geoscience and Remote Sensing}, vol.~56, no.~2, pp. 937--949,
  2017.

\bibitem{cnn3d}
A.~B. Hamida, A.~Benoit, P.~Lambert, and C.~B. Amar, ``3-d deep learning
  approach for remote sensing image classification,'' \emph{IEEE Transactions
  on Geoscience and Remote Sensing}, vol.~56, no.~8, pp. 4420--4434, 2018.

\bibitem{luo2018hsi}
Y.~Luo, J.~Zou, C.~Yao, X.~Zhao, T.~Li, and G.~Bai, ``Hsi-cnn: A novel
  convolution neural network for hyperspectral image,'' in \emph{2018
  International Conference on Audio, Language and Image Processing
  (ICALIP)}.\hskip 1em plus 0.5em minus 0.4em\relax IEEE, 2018, pp. 464--469.

\bibitem{roy2019hybridsn}
S.~K. Roy, G.~Krishna, S.~R. Dubey, and B.~B. Chaudhuri, ``Hybridsn: Exploring
  3-d--2-d cnn feature hierarchy for hyperspectral image classification,''
  \emph{IEEE Geoscience and Remote Sensing Letters}, vol.~17, no.~2, pp.
  277--281, 2019.

\bibitem{zhong2017spectral}
Z.~Zhong, J.~Li, Z.~Luo, and M.~Chapman, ``Spectral--spatial residual network
  for hyperspectral image classification: A 3-d deep learning framework,''
  \emph{IEEE Transactions on Geoscience and Remote Sensing}, vol.~56, no.~2,
  pp. 847--858, 2017.

\bibitem{paoletti2018deep}
M.~E. Paoletti, J.~M. Haut, R.~Fernandez-Beltran, J.~Plaza, A.~J. Plaza, and
  F.~Pla, ``Deep pyramidal residual networks for spectral-spatial hyperspectral
  image classification,'' \emph{IEEE Transactions on Geoscience and Remote
  Sensing}, vol.~57, no.~2, pp. 740--754, 2018.

\bibitem{he2021spatial}
X.~He, Y.~Chen, and Z.~Lin, ``Spatial-spectral transformer for hyperspectral
  image classification,'' \emph{Remote Sensing}, vol.~13, no.~3, p. 498, 2021.

\bibitem{dosovitskiy2020image}
A.~Dosovitskiy, L.~Beyer, A.~Kolesnikov, D.~Weissenborn, X.~Zhai,
  T.~Unterthiner, M.~Dehghani, M.~Minderer, G.~Heigold, S.~Gelly \emph{et~al.},
  ``An image is worth 16x16 words: Transformers for image recognition at
  scale,'' \emph{arXiv preprint arXiv:2010.11929}, 2020.

\bibitem{zhang2022convolution}
J.~Zhang, Z.~Meng, F.~Zhao, H.~Liu, and Z.~Chang, ``Convolution transformer
  mixer for hyperspectral image classification,'' \emph{IEEE Geoscience and
  Remote Sensing Letters}, vol.~19, pp. 1--5, 2022.

\bibitem{sun2022spectral}
L.~Sun, G.~Zhao, Y.~Zheng, and Z.~Wu, ``Spectral--spatial feature tokenization
  transformer for hyperspectral image classification,'' \emph{IEEE Transactions
  on Geoscience and Remote Sensing}, vol.~60, pp. 1--14, 2022.

\bibitem{mei2022hyperspectral}
S.~Mei, C.~Song, M.~Ma, and F.~Xu, ``Hyperspectral image classification using
  group-aware hierarchical transformer,'' \emph{IEEE Transactions on Geoscience
  and Remote Sensing}, vol.~60, pp. 1--14, 2022.

\bibitem{hong2021spectralformer}
D.~Hong, Z.~Han, J.~Yao, L.~Gao, B.~Zhang, A.~Plaza, and J.~Chanussot,
  ``Spectralformer: Rethinking hyperspectral image classification with
  transformers,'' \emph{IEEE Transactions on Geoscience and Remote Sensing},
  vol.~60, pp. 1--15, 2021.

\bibitem{zhong2021spectral}
Z.~Zhong, Y.~Li, L.~Ma, J.~Li, and W.-S. Zheng, ``Spectral--spatial transformer
  network for hyperspectral image classification: A factorized architecture
  search framework,'' \emph{IEEE Transactions on Geoscience and Remote
  Sensing}, vol.~60, pp. 1--15, 2021.

\bibitem{morphformer}
S.~K. Roy, A.~Deria, C.~Shah, J.~M. Haut, Q.~Du, and A.~Plaza,
  ``Spectral–spatial morphological attention transformer for hyperspectral
  image classification,'' \emph{IEEE Transactions on Geoscience and Remote
  Sensing}, vol.~61, pp. 1--15, 2023.

\bibitem{hetconv}
P.~Singh, V.~K. Verma, P.~Rai, and V.~P. Namboodiri, ``Hetconv: Heterogeneous
  kernel-based convolutions for deep cnns,'' 2019.

\bibitem{batchnormalization}
S.~Ioffe and C.~Szegedy, ``Batch normalization: Accelerating deep network
  training by reducing internal covariate shift,'' 2015.

\bibitem{visionTransformer}
A.~Dosovitskiy, L.~Beyer, A.~Kolesnikov, D.~Weissenborn, X.~Zhai,
  T.~Unterthiner, M.~Dehghani, M.~Minderer, G.~Heigold, S.~Gelly, J.~Uszkoreit,
  and N.~Houlsby, ``An image is worth 16x16 words: Transformers for image
  recognition at scale,'' 2021.

\bibitem{ahmad2021hyperspectral}
M.~Ahmad, S.~Shabbir, S.~K. Roy, D.~Hong, X.~Wu, J.~Yao, A.~M. Khan,
  M.~Mazzara, S.~Distefano, and J.~Chanussot, ``Hyperspectral image
  classification—traditional to deep models: A survey for future prospects,''
  \emph{IEEE Journal of Selected Topics in Applied Earth Observations and
  Remote Sensing}, vol.~15, pp. 968--999, 2021.

\bibitem{svm}
F.~Melgani and L.~Bruzzone, ``Classification of hyperspectral remote sensing
  images with support vector machines,'' \emph{IEEE Transactions on Geoscience
  and Remote Sensing}, 2004.

\bibitem{rf}
M.~Ahmad, S.~Shabbir, S.~K. Roy, D.~Hong, X.~Wu, J.~Yao, A.~M. Khan,
  M.~Mazzara, S.~Distefano, and J.~Chanussot, ``Hyperspectral image
  classification—traditional to deep models: A survey for future prospects,''
  \emph{IEEE Journal of Selected Topics in Applied Earth Observations and
  Remote Sensing}, vol.~15, pp. 968--999, 2021.

\bibitem{rnn}
L.~Mou, P.~Ghamisi, and X.~X. Zhu, ``Deep recurrent neural networks for
  hyperspectral image classification,'' \emph{IEEE Transactions on Geoscience
  and Remote Sensing}, vol.~55, no.~7, pp. 3639--3655, 2017.

\end{thebibliography}
}

\end{document}